%% file: StreamTraj.tex
\documentclass{article}

\usepackage[final]{neurips_2023}

\usepackage[utf8]{inputenc} % allow utf-8 input
\usepackage[T1]{fontenc}    % use 8-bit T1 fonts
\usepackage{hyperref}       % hyperlinks
\usepackage{booktabs}       % professional-quality tables
\usepackage{amsfonts}       % blackboard math symbols
\usepackage{nicefrac}       % compact symbols for 1/2, etc.
\usepackage{microtype}      % microtypography

\usepackage[american]{babel}
\usepackage{natbib} % has a nice set of citation styles and commands
\usepackage{mathtools} % amsmath with fixes and additions

\usepackage{tikz} % nice language for creating drawings and diagrams

\usepackage{graphicx}
\usepackage{natbib}
\usepackage{amssymb, amsmath,amsthm}
\usepackage{algorithm}
\usepackage{algorithmic}
\usepackage{paralist}
\usepackage{multirow}
\usepackage{times}
\usepackage{wrapfig}
\usepackage{url,enumerate}
\usepackage{color,xcolor}
\usepackage{makeidx}  % allows for indexgeneration
\usepackage{amsmath,amssymb}
\usepackage{mathtools}
\usepackage{xspace}
\usepackage{epstopdf}
\usepackage{mathrsfs}
\usepackage{times}
\usepackage{enumerate}
\usepackage{enumitem}

\usepackage{color}
\usepackage{epsfig}
\usepackage{amsmath,amssymb,xspace}
\usepackage{bm}
\usepackage{bbm}
\usepackage{upgreek}
\usepackage{cleveref}
\usepackage{multirow}
\usepackage{ulem}
\usepackage{cancel}
\usepackage{subcaption}
\usepackage{hyperref}

\newcommand{\ours}{{{SFTL}}\xspace}

\input{emacscomm.tex}

\title{Streaming Factor Trajectory Learning for Temporal Tensor Decomposition}

\author{%
	Shikai Fang\\
	Kahlert School of Computing\\
	The University of Utah\\
	\texttt{shikai.fang@utah.edu} \\
	\And
	Xin Yu \\
	Kahlert School of Computing\\
	The University of Utah\\ 
	\texttt{yuxwind@gmail.com} \\
	\And
	Shibo Li \\
	Kahlert School of Computing\\
	The University of Utah\\ 
	\texttt{shiboli.cs@gmail.com} \\
	\And
	Zheng Wang \\
	Kahlert School of Computing\\
	The University of Utah\\ 
	\texttt{u1208847@utah.edu} \\
	\And
	Robert M. Kirby \\
	Kahlert School of Computing\\
	The University of Utah\\ 
	\texttt{kirby@cs.utah.edu} \\
	\And
	Shandian Zhe\thanks{Corresponding author.} \\
	Kahlert School of Computing\\
	The University of Utah\\ 
	\texttt{zhe@cs.utah.edu} \\
}

\begin{document}

\maketitle
\input{./abstract}
\input{./intro}

%\input{./Background}
\input{./method}
% \input{./method-fang}

\input{./related}
\input{./exp-zhe}
\input{./conclusion}

\bibliographystyle{apalike}
\bibliography{StreamTraj}

\newpage
\appendix

\input{./supp-zhe}

\end{document}

%% file: emacscomm.tex
\newcommand{\kron}{\otimes}

\newcommand{\diag}{{\rm diag}}

\renewcommand{\vec}{{\rm vec}}

\renewcommand{\d}{{\rm d}}  % for derivatives

\newcommand{\f}{{\bf f}}
\newcommand{\g}{{\bf g}}
\newcommand{\h}{{\bf h}}
\newcommand{\m}{{\bf m}}

\renewcommand{\r}{{\bf r}}

\renewcommand{\u}{{\bf u}}
\renewcommand{\v}{{\bf v}}

\newcommand{\x}{{\bf x}}
\newcommand{\y}{{\bf y}}
\newcommand{\z}{{\bf z}}
\newcommand{\A}{{\bf A}}

\newcommand{\F}{{\bf F}}
\newcommand{\G}{{\bf G}}
\renewcommand{\H}{{\bf H}}
\newcommand{\I}{{\bf I}}
\newcommand{\J}{{\bf J}}
\newcommand{\K}{{\bf K}}

\newcommand{\N}{\mathcal{N}}  % for normal density

\newcommand{\Bcal}{\mathcal{B}}

\newcommand{\Dcal}{\mathcal{D}}
\newcommand{\Ocal}{\mathcal{O}}
\newcommand{\Ycal}{\mathcal{Y}}
\newcommand{\Ical}{\mathcal{I}}

\newcommand{\Vcal}{\mathcal{V}}

\newcommand{\gp}{\mathcal{GP}}

\renewcommand{\P}{{\bf P}}
\newcommand{\Q}{{\bf Q}}

\renewcommand{\S}{{\bf S}}

\newcommand{\V}{{\bf V}}
\newcommand{\W}{{\bf W}}

\newcommand{\Wcal}{{\mathcal{W}}}
\newcommand{\Ucal}{{\mathcal{U}}}

\newcommand{\bphi}{\boldsymbol{\phi}}
\newcommand{\boldeta}{\boldsymbol{\eta}}
\newcommand{\bbeta}{\boldsymbol{\beta}}

\newcommand{\blambda}{\boldsymbol{\lambda}}

\newcommand{\bOmega}{\mathbf{\Omega}}

\newcommand{\btheta}{\boldsymbol{\theta}}

\newcommand{\bSigma}{\boldsymbol{\Sigma}}

\newcommand{\bgamma}{\boldsymbol{\gamma}}

\newcommand{\bell}{{\boldsymbol \ell}}
\newcommand{\barz}{{\h}}

\newcommand{\bmu}{\boldsymbol{\mu}}
\newcommand{\1}{{\bf 1}}
\newcommand{\0}{{\bf 0}}

\newcommand{\ben}{\begin{enumerate}}
\newcommand{\een}{\end{enumerate}}

\newcommand{\argmin}{\operatornamewithlimits{argmin}}

\newcommand{\ie}{{\textit{i.e.,}}\xspace}
\newcommand{\eg}{{\textit{e.g.,}}\xspace}
\newcommand{\etc}{{\textit{etc.}}\xspace}
\newcommand{\EE}{\mathbb{E}}

\newcommand{\cmt}[1]{}

\newcommand{\br}{{\bf r}}

\newcommand{\tp}{{\widetilde{p}}}
\newcommand{\talpha}{{\widetilde{\alpha}}}
\newcommand{\tomega}{{\widetilde{\omega}}}
\newcommand{\bkh}{{\backslash}}
\newcommand{\kl}{{\mathrm{KL} }}

\newcommand{\appropto}{\mathrel{\vcenter{
			\offinterlineskip\halign{\hfil$##$\cr
				\propto\cr\noalign{\kern2pt}\sim\cr\noalign{\kern-2pt}}}}}

%% file: abstract.tex
%continuous time issue --- our method, interpretability & nonparametric --- 
\begin{abstract}
Practical tensor data is often along with time information. Most existing temporal decomposition approaches estimate a set of fixed factors for the objects in each tensor mode, and hence cannot capture the temporal evolution of the objects' representation. More important, we lack an effective approach to capture such evolution from streaming data, which is common in real-world applications.  To address these issues, we propose Streaming Factor Trajectory Learning (\ours) for temporal tensor decomposition. We use Gaussian processes (GPs) to model the trajectory of  factors so as to flexibly estimate their temporal evolution. To address the computational challenges in handling streaming data, we convert the GPs into a state-space prior by constructing an equivalent stochastic differential equation (SDE).  We develop an efficient online filtering algorithm to estimate a decoupled running posterior of the involved factor states upon receiving new data. The decoupled estimation enables us to conduct standard Rauch-Tung-Striebel smoothing to compute the full posterior of all the  trajectories in parallel, without the need for revisiting any previous data. We have shown the advantage of \ours in both synthetic tasks and real-world applications.  The code is available at \url{https://github.com/xuangu-fang/Streaming-Factor-Trajectory-Learning}.
\end{abstract}
%Practical tensor data is often along with time information. Most existing temporal decomposition approaches estimate a set of fixed factors for the objects in each tensor mode. Therefore, these methods cannot capture the temporal evolution of the objects' representation. More important, we lack an effective approach to capture such evolution from streaming data, which is ubiquitous in real-world applications.  To address these issues, we propose Streaming Factor Trajectory Learning for Temporal Tensor Decomposition (\ours). We use Gaussian processes (GPs) to model a trajectory function of the factors so as to flexibly estimate their temporal evolution. To address the computational challenges in handling streaming data, we convert the GPs into a state-space prior by constructing an equivalent stochastic differential equation (SDE). 
%We use Kalman-filtering and moment matching to instantly compute the running posterior of the factor states upon receiving new tensor elements. After all the data is processed, we conduct one-pass RTS smoothing to efficiently compute the posterior of the full trajectory, without the need for revisiting any previous data. We have shown the advantage of our method in both synthetic data and real-world applications. 

%% file: intro.tex
\section{Introduction}
%tensor decomposition background
Tensor data is common in real-world applications. For example, one can extract a three-mode tensor \textit{(patient, drug, clinic)} from medical service records and  a four-mode tensor \textit{(customer, commodity, seller, web-page)} from the database of an online shopping platform. Tensor decomposition is a fundamental tool for tensor data analysis. It introduces a set of factors to represent the objects in each mode, and estimate these factors by reconstructing the observed entry values. These factors can be viewed as the underlying properties of the objects. We can use them to search for interesting structures within the objects (\eg communities and outliers) or as discriminate features for predictive tasks, such as personalized treatment or recommendation.

Real-world tensor data is often accompanied with time information, namely the timestamps at which the objects of different modes interact to produce the entry values. Underlying the timestamps can be rich, valuable temporal patterns. While many temporal decomposition methods are available, most of them estimate a set of static factors for each object --- they either introduce a discrete time mode~\citep{xiong2010temporal,zhe2016dintucker} or inject the timestamp into the decomposition model~\citep{zhang2021dynamic,fang2022bayesian,li2022decomposing}. Hence, these methods cannot learn the temporal variation of the factors. Accordingly, they can miss important evolution of the objects' inner properties, such as health and income.  In addition, practical applications produce data streams  at 
a rapid pace~\citep{du2018probabilistic}. Due to the resource limit, it is often  prohibitively  expensive to decompose the entire tensor from scratch whenever we receive new data. Many privacy-protecting applications (\eg SnapChat) even forbid us to preserve or re-access  the previous data. Therefore,  not only do we need a more powerful decomposition model that can estimate the factor evolution, we also need an effective method to capture such evolution from fast data streams.

 To address these issues, we propose \ours, a Bayesian streaming factor trajectory learning approach for temporal tensor decomposition. Our method can efficiently handle data streams, with which to estimate a posterior distribution of the factor  trajectory to uncover the temporal evolution of the objects' representation. Our method never needs to keep or re-visit previous data. The contribution of our work is summarized  as follows. 
 \begin{compactitem}
 	\item First, we use a Gaussian process (GP) prior to sample the factor trajectory of each object as a function of time. As a nonparametric function prior, GPs are flexible enough to capture a variety of complex temporal dynamics.   The trajectories are combined through a  CANDECOMP/PARAFAC (CP)~\citep{Harshman70parafac} or Tucker decomposition~\citep{Tucker66} form to sample the tensor entry values at any time point. %In this way, our model also maintains the interpretability of the classical decomposition models. 
 	\item Second, to sidestep the expensive covariance matrix computation in the GP, which further causes challenges in streaming inference, we use spectral analysis to convert the GP into a linear time invariant (LTI) stochastic differential equation (SDE). Then we convert the SDE into  an equivalent  state-space prior over the factor (trajectory) states at the observed timestamps.   As a result, the posterior inference becomes  easier and computationally efficient.
 	\item Third, we take advantage of the chain structure of the state-space prior and use the recent conditional expectation propagation framework~\citep{wang2019conditional} to develop an efficient online filtering algorithm. Whenever a collection of entries at a new timestamp arrives, our method can efficiently estimate a decoupled running posterior of the involved factor states, with a Gaussian product form. The decoupled Gaussian estimate enables us to run standard RTS smoothing~\citep{sarkka2013bayesian} to compute the full posterior of each factor trajectory independently and in parallel, without revisiting any previous data. Our method at worst has a linear scalability with respect to the number of observed timestamps. 
 \end{compactitem}

We first evaluated our method in a simulation study. On synthetic datasets, \ours successfully recovered several nonlinear factor trajectories, and provided reasonable uncertainty estimation. We then tested our method on four real-world temporal tensor datasets for missing value prediction.  In both online and final predictive
performance,  \ours consistently outperforms the state-of-the-art streaming CP and Tucker decomposition algorithms by a large margin. In most cases, the prediction accuracy of \ours is even higher than the recent static decomposition methods, which have to pass through the dataset many times. Finally, we investigated the learned factor trajectories from a real-world dataset. The trajectories exhibit interesting temporal evolution.

%% file: method.tex
%background: CP/Tucker/GP decompostion

%method: model --> spectral analysis to derive state-space prior --> inference algorithm
\section{Preliminaries}\label{sect:bk}
\textbf{Tensor Decomposition.} We denote an $M$-mode tensor by $\Ycal \in \mathbb{R}^{d_1 \times \cdots \times d_M}$, where each mode $m$ has $d_m$ dimensions, corresponding to $d_m$ objects. Each tensor entry is indexed by a tuple $\bell = \left(\ell_1, \ldots, \ell_M\right)$, and the value is denoted by $y_\bell$. For decomposition, we introduce a set of latent factors $\u^m_j \in \mathbb{R}^{R_m}$ to represent each object $j$ in mode $m$ ($1 \le m \le M$). One most popular tensor decomposition model is the CANDECOMP/PARAFAC (CP) decomposition~\citep{Harshman70parafac}, which sets $R_1 = \ldots = R_M = R$, and uses the following element-wise form, $y_{\bell} \approx \1^\top (\u^1_{\ell_1} \circ \ldots \circ \u^M_{\ell_M})= \sum\nolimits_{r=1}^{R}  \prod\nolimits_{m=1}^{M} u_{\ell_{m}, r}^{m}$,
where $\circ$ is the element-wise product. 
Another commonly used model is Tucker decomposition~\citep{Tucker66}, $y_\bell \approx \vec(\Wcal )^{\top}\left(\u_{\ell_{1}}^{1} \otimes \ldots \otimes \u_{\ell_{M}}^{M}\right) =\sum\nolimits_{r_{1}=1}^{R_{1}} \ldots \sum\nolimits_{r_{M}=1}^{R_{M}}\left[w_{\br} \cdot \prod\nolimits_{m=1}^{M} u_{\ell_{m}, r_{m}}^{m}\right]$, 
where $\Wcal \in \mathbb{R}^{R_1 \times \cdots \times R_M} $ is the tensor-core parameter, $\vec(\cdot)$ is the vectorization, $\otimes$ is the Kronecker product,  and $\r = (r_1, \ldots, r_M)$.

\textbf{Gaussian Process (GP)s} are  nonparametric function priors. For a function $f(\x)$, if we place a GP prior, $f \sim \gp(0, \kappa(\x, \x'))$, it means $f(\cdot)$ is sampled as a realization of the GP with covariance function $\kappa$, which is often chosen as a kernel function. The GP prior only models the correlation between the function values, namely, $\text{cov}\left(f(\x), f(\x')\right) = \kappa(\x, \x')$, and does not assume any parametric form of the function. Hence, GPs are flexible enough to estimate various complex functions from data, \eg from multilinear to highly nonlinear. The finite projection of the GP is a Gaussian distribution. That is, given an arbitrary collection of inputs $\{\x_1, \ldots, \x_N\}$, the corresponding function values $\f = [f(\x_1), \ldots, f(\x_N)]^\top$ follow a multi-variate Gaussian prior distribution, $p(\f) = \N(\f|\0, \K)$ where $\K$ is the covariance matrix and each $[\K]_{mn} = \kappa(\x_m, \x_n)$.

%way up to consder the modeo joint probability --> inference hurdle --> state space prior --> new probability
\section{Bayesian Temporal Tensor Decomposition with Factor Trajectories}
In real-world applications, tensor data is often associated with time information, namely, the timestamps at which the objects of different modes interact to generate the entry values. 
To capture the potential evolution of the objects' inner properties, we propose a Bayesian temporal tensor decomposition model that can estimate a trajectory of the factor representation. Specifically, for each object $j$ in mode $m$, we model the factors as a function of time, $\u^m_j : [0, \infty] \rightarrow \mathbb{R}^R$. To flexibly capture a variety of temporal evolution, we assign a GP prior over each element of $\u^m_j(t) = [u^m_{j,1}(t), \ldots, u^m_{j,R}(t)]^\top$,  \ie $u^m_{j,r}(t) \sim \gp(0, \kappa(t, t'))$ ($1 \le r \le R$). 
%\begin{align}
	%u^m_{jr}(t) \sim \gp(0, \kappa(t, t')) \label{eq:gp-u}
%\end{align}
%where $1 \le r \le R$. 
Given the factor trajectories, we then use the CP or Tucker form to sample the entry values at different time points. For the CP form, we have  
\begin{align}
	&p(y_\bell(t)|\Ucal(t)) = \N(y_\bell(t) | \1^\top (\u^1_{\ell_1}(t) \circ \ldots \circ \u^M_{\ell_M}(t)), \tau^{-1}), \label{eq:cp-ll}
\end{align}
where $\Ucal(t) = \{\u^m_{j}(t)\}$ includes all the factor trajectories, and 
$\tau$ is the inverse noise variance, for which we assign a Gamma prior, $p(\tau) = \text{Gam}(\tau|\alpha_0, \alpha_1)$. For the Tucker form, we have $p(y_\bell(t)|\Ucal(t), \Wcal) = \N(y_\bell(t) | \vec(\Wcal)^\top (\u^1_{\ell_1}(t) \kron \ldots \kron \u^M_{\ell_M}(t)), \tau^{-1})$
%\begin{align}
%	&p(y_\bell(t)|\Ucal(t), \Wcal) \label{eq:tucker-ll}  \\
%	&= \N(y_\bell(t) | \vec(\Wcal)^\top (\u^1_{\ell_1}(t) \kron \ldots \kron \u^K_{\ell_M}(t)), \tau^{-1}), \notag
%\end{align}
where we place a standard normal prior over the tensor-core, $p(\vec(\Wcal)) = \N(\vec(\Wcal)|\0, \I)$.  In this work, we focus on continuous observations. It is straightforward to extend our method for  other types of observations. 

Suppose we have a collection of observed entry values and timestamps, $\Dcal = \{(\bell_1, y_1, t_1), \ldots, (\bell_N, y_N, t_N)\}$ where $t_1 \le \cdots \le t_N$.  We denote the sequence of timestamps when a particular object $j$ of mode $m$ participated in the observed entries by $s^m_{j,1} < \ldots < s^m_{j,c^m_j}$, where $c^m_j$ is the participation count of the object. Note that it is a sub-sequence of $\{t_n\}$. From the GP prior, the values of each $u^m_{j,r}(t)$ at these timestamps follow a multi-variate Gaussian distribution, $p(\u^m_{j,r}) = \N(\u^m_{j,r}|\0, \K^m_{j})$ where $\u^m_{j,r} = [u^m_{j,r}(s^m_{j,1}), \ldots, u^m_{j,r}(s^m_{j,c^m_j}) ]^\top$\footnote{For convenience, we abuse the notation a little bit: we denote by $\u^m_{j,r}(\cdot)$ trajectory function and by $\u^m_{j,r}$ the values of the trajectory function at the observed timestamps.} and $\K^m_j$ is the covariance/kernel matrix computed at these timestamps. The joint probability with the CP form is  
\begin{align}
	&p(\{\u^m_{j,r}\},\tau, \y) =   \prod\nolimits_{m=1}^M \prod\nolimits_{j=1}^{d_m} \prod\nolimits_{r=1}^R \N(\u^m_{j,r}|\0, \K^m_j) \cdot \text{Gam}(\tau|\alpha_0, \alpha_1)\notag \\
	&  \cdot \prod\nolimits_{n=1}^N \N(y_n | \1^\top (\u^1_{\ell_{n1}}(t_n) \circ \ldots \circ \u^M_{\ell_{nM}}(t_n)), \tau^{-1}). \label{eq:joint-1} \raisetag{0.2in}
\end{align}
The joint probability with the Tucker form is the same  except that we use the Tucker likelihood instead and multiply with the prior of  tensor-core $p(\Wcal)$.

While this formulation is straightforward, it can introduce computational challenges. There are many multi-variate Gaussian distributions in the joint distribution \eqref{eq:joint-1}, \ie $\{\N(\u^m_{j,r}|\0, \K^m_j)\}$. The time and space complexity to compute each $\N(\u^m_{j,r}|\0, \K^m_j)$ is $\Ocal\left(\left(c^m_j\right)^3\right)$ and $\Ocal\left(\left(c^m_j\right)^2\right)$, respectively. With the increase of $N$, the appearance count $c^m_j$ for many objects can grow as well, making the computation cost very expensive or even infeasible. The issue is particularly severe when we handle streaming data --- the number of timestamps grows rapidly when new data keeps coming in, so does the size of each covariance matrix. 
\subsection{Equivalent Modeling with State-Space Priors}
To sidestep expensive covariance matrix computation and ease the inference with streaming data, we follow~\citep{hartikainen2010kalman} to convert the GP prior into an SDE via spectral analysis. We use a Mat\'ern kernel $\kappa_\nu\left(t, t'\right) = a \frac{\left(\frac{\sqrt{2 \nu}}{\rho} \Delta \right)^{\nu}}{\Gamma(\nu) 2^{\nu-1}}
K_{\nu}\left(\frac{\sqrt{2 \nu}}{\rho} \Delta\right) $ where $\Gamma(\cdot)$ is the Gamma function, $\Delta = |t-t'|$, $a>0$, $\rho>0$, $K_{\nu}$ is the modified Bessel function of the second kind, and $\nu = p + \frac{1}{2}$ ($p\in \{0, 1, 2, \ldots\}$) as the GP covariance. Via the analysis of the power spectrum of $\kappa_\nu$, we can show that  if $f(t) \sim \gp(0, \kappa_\nu(t, t'))$, it can be characterized by a linear time-invariant (LTI) SDE, with  state $\z = (f, f^{(1)}, \ldots, f^{(p)})^\top$ where $f^{(k)} \overset{\Delta}{=} \d^k f/\d t^k$,  
\begin{align}
	\frac{\d \z}{\d t} = \A \z + \boldeta \cdot \beta(t),
\end{align}
where $\beta(t)$ is a white noise process with diffusion $\sigma^2$, 
\begin{align}
	\A =\left(\begin{array}{cccc}
		0 & 1 & & \\
		& \ddots  & \ddots & \\
		&   & 0 & 1 \\
		-c_0 & \ldots & -c_{p-1} & -c_{p}
	\end{array}\right), \quad \boldeta =\left(\begin{array}{c}
		0 \\
		\vdots\\
		0 \\
		1\\
	\end{array}\right). \notag 
\end{align}
Both $\sigma^2$ and $\A$ are obtained from the parameters in $\kappa_\nu$. Due to the space limit, we leave the detailed derivation in Appendix (Section \ref{appendix:sect:A}). The LTI-SDE is particularly useful  in that its finite set of states follow a Gauss-Markov chain, \ie the state-space prior.  Given arbitrary $t_1 < \ldots < t_L$, we have 
\begin{align}
	p(\z(t_1), \ldots, \z(t_L)) = p(\z(t_1)) \prod\nolimits_{k=1}^{L-1} p(\z(t_{k+1})| \z(t_{k})), \notag 
\end{align}
where $p(\z(t_1)) = \N(\z(t_1)|\0, \P_\infty)$, $p(\z(t_{k+1})| \z(t_{k})) = \N(\z(t_{k+1}) |\F_{k} \z(t_{k}), \Q_k)$, 
%\begin{align}
%	p(\z(t_1)) &= \N(\z(t_1)|\0, \P_\infty), \notag \\
%	p(\z(t_{k+1})| \z(t_{k})) &= \N(\z(t_{k+1}) |\F_{k} \z(t_{k}), \Q_k)
%\end{align}
$\P_\infty$ is the stationary covariance matrix computed by solving the matrix Riccati equation~\citep{lancaster1995algebraic}\cmt{ (\eg $\P_\infty = [a, 0; 0, a\alpha^2]$ when $n=1$)}, $\F_n = \exp(\Delta_k \cdot \A )$ where $\Delta_k = t_{k+1} - t_k$, and $\Q_k = \P_\infty - \A_k \P_\infty \A_k^\top$.  Therefore, we  do not  need the full covariance matrix as in the standard GP prior, and the computation is much more efficient. The chain structure is also convenient to handle streaming data as we will explain later. %Note that for other type of kernel functions, such as the square exponential (SE) kernel, we can approximate the inverse spectral density $1/S(\omega)$ with a polynomial of $\omega^2$ with negative roots, and follow the same way to construct an LTI-SDE and state-space prior. 

We therefore convert the GP prior over each factor trajectory $u^m_{j,r}(t)$ into an LTI-SDE. We denote the corresponding state by $\z^m_{j,r}(t)$. For example, if we choose $p=1$, then $\z^m_{j,r}(t) = [u^m_{j,r}(t); \d u^m_{j,r}(t)/\d t]$. 
 For each object $j$ in mode $m$, we concatenate all its trajectory states into one,  $\z^m_j(t) = [\z^m_{j,1}(t); \ldots; \z^m_{j,R}(t)]$. Then on all of its timestamps $s^m_{j,1} < \ldots < s^m_{j,c^m_j}$, we obtain a state-space prior 
\begin{align}
	p(\barz^m_{j,1}) = \N(\barz^m_{j,1} | \0, \overline{\P}_\infty), \;\;	p(\barz^m_{j,k+1} | \barz^m_{j,k}) = \N(\barz^m_{j,k+1}| \overline{\F}^m_{j,k}  \barz^m_{j,k}, \overline{\Q}^m_{j,k}), \label{eq:state-space}
\end{align}
where $\barz^m_{j,k} \overset{\Delta}{=} \z^m_j(s^m_{j,k})$, 
$\overline{\P}_\infty = \diag(\P_\infty, \ldots, \P_\infty)$, $\overline{\F}^m_{j,k} = \diag\left(\F^m_{j,k}, \ldots, \F^m_{j,k}\right)$, $\F^m_{j,k} = e^{ (s^m_{j,k+1} - s^m_{j,k})\A}$,  $\overline{\Q}^m_{j,k} = \diag\left(\Q^m_{j,k}, \ldots, \Q^m_{j,k}\right)$, and $\Q^m_{j,k}  = \P_\infty - \F^m_{j,k} \P_\infty \left(\F^m_{j,k}\right)^\top$. 

 The joint probability of our model with the CP form now becomes 
\begin{align}
	&p(\{\barz^m_{j,k}\}, \tau, \y)  =p(\tau) \prod\nolimits_{m=1}^M \prod\nolimits_{j=1}^{d_m} p(\barz^m_{j,1}) \prod\nolimits_{k=1}^{c^m_j - 1} 	p(\barz^m_{j,k+1} | \barz^m_{j,k}) \notag \\
	%\text{Gam}(\tau|a_0, b_0)\notag \\
	& \cdot  \prod\nolimits_{n=1}^N \N(y_n | \1^\top (\u^1_{\ell_{n1}}(t_n) \circ \ldots \circ \u^M_{\ell_{nM}}(t_n)), \tau^{-1}). \label{eq:new-prob} \raisetag{0.22in}
\end{align}
Note that in the likelihood, each $\u^m_{{\ell_{nm}}}(t_n) (1 \le j \le M)$ is contained in a corresponding state vector $\barz^m_{{\ell_{nm}}, k}$ such that $s^m_{\ell_{nm}, k} = t_n$ (by definition, we then have $\barz^m_{{\ell_{nm}}, k} = \z^m_{\ell_{nm}}(t_n)$). The joint probability with the Tucker form is similar, which we omit to save the space. 
\begin{figure*}
	\centering
	\setlength\tabcolsep{0pt}
	\includegraphics[width=0.8\linewidth]{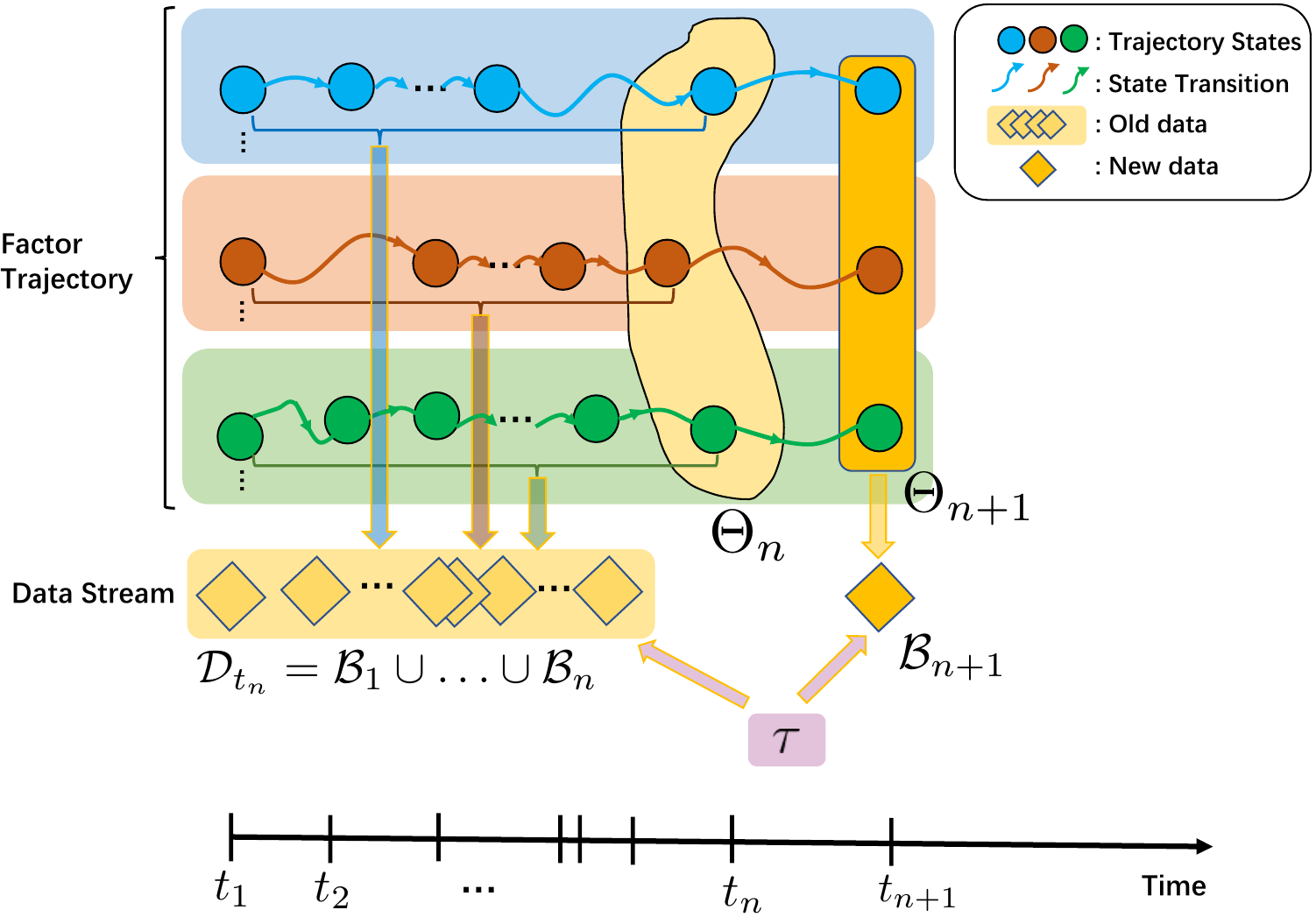}
	\caption{\small  A graphical representation of our factor trajectory learning, from which we can see $\{\Theta_{n+1}, \Bcal_{n+1}\}$ are independent to $\Dcal_{t_{n}}$ conditioned on $\Theta_{n}$ and the noise inverse variance $\tau$, namely, $\Theta_{n+1}, \Bcal_{n+1} \perp 
		\Dcal_{t_n} | \Theta_n, \tau$.}
	\label{fig:graphical_model}
\end{figure*}
\section{Trajectory Inference from Streaming Data}
%problem definition
In this section, we develop an efficient, scalable algorithm for factor trajectory estimation from streaming data. In general, we assume that we receive a sequence of (small) batches of observed tensor entries, $\{\Bcal_1, \Bcal_2, \ldots\}$, generated at different timestamps, $\{t_1, t_2, \ldots\}$. Each batch $\Bcal_n$ is generated at timestamp $t_n$ and $t_n < t_{n+1}$. \cmt{Note that the same entry can be observed as different values at multiple timestamps.}  Denote by  $\Dcal_{t_n}$ all the data up to timestamp $t_n$, \ie $\Dcal_{t_n} = \Bcal_1 \cup \ldots \cup \Bcal_{n}$.  Upon receiving  $\Bcal_{n+1}$\cmt{ --- a batch of newly observed entries at $t_{n+1}$},  we intend to update our model without revisiting $\Dcal_{t_n}$ to provide the trajectory posterior estimate, $\{p(\u^m_j(t) | \Dcal_{t_{n+1}})| \forall t\ge 0, 1 \le m \le M, 1 \le j \le d_m\}$, where $\Dcal_{t_{n+1}} = \Dcal_{t_{n}} \cup \Bcal_{n+1}$. %Accordingly, we are  We are able to c compute the predictive distribution for an arbitrary entry value at an arbitrary time point $t$. 

%our observation, and RTS 
To this end, we first observe that the standard state-space model with a Gaussian likelihood has already provided a highly efficient streaming inference framework. Denote by $\x_n$ and $\y_n$ the state and observation at each step $n$, respectively. To handle streaming observations $\{ \y_1, \y_2, \ldots \}$, we only need to compute and track the running posterior $p(\x_n | \y_{1:n})$  upon receiving each $\y_n$, where $\y_{1:n} $ denotes the total data up to step $n$.  This is called Kalman filtering~\citep{sarkka2013bayesian}, which only depends on the running posterior at step $n-1$, \ie $p(\x_{n-1} | \y_{1:n-1})$, and is highly efficient. After all the data is processed (suppose it stops at step $N$),  we can use Rauch-Tung-Striebel (RTS) smoother~\citep{sarkka2013bayesian} to efficiently compute the full posterior of each state, $p(\x_n|\y_{1:N})$,  from backward, which does not need to re-access any previous observations (see Section \ref{sect:rts} in Appendix).

However, one cannot apply  the above framework  outright to our model, since the tensor decomposition likelihood of each observed entry couples the states of multiple factor trajectories, see \eqref{eq:cp-ll} --- which correspond to the participated objects at different modes. That means, the factor-state chains of different objects are dynamically intertwined through the received data. The multiplicative form of these states in the likelihood render the running posterior of each trajectory intractable to compute, not to mention running RTS smoother. To address this challenge, we take advantage of the chain structure and use the recent conditional Expectation propagation (CEP) framework~\citep{wang2019conditional} to develop an efficient online filtering algorithm, which approximates  the running posterior of the involved factor states as a product of Gaussian. %each time upon receiving new data. %Due to the factorized structure in the likelihood approximation, the running posterior estimation will also be factorized over each factor state. 
Thereby, we can decouple the involved factor state chains, and conduct standard RTS smoothing for each chain independently. 

%notations, marginalize out ..., and give the formulation, and refer to Zheng's paper --- it is done 
Specifically, denote the sequence of timestamps  when each object $j$ of mode $m$ has showed up in the data stream up to $t_n$, by $s^m_{j, 1} < s^m_{j,2}<\ldots < s^m_{j, c^m_{j,n}}$ where $c^m_{j,n}$ is the object's appearance count up to $t_n$.  \cmt{This is a sub-sequence of $t_1 < \cdots < t_n$. } 
 Denote by $\Ical^m_{n}$ the indexes of all the objects of mode $m$ appearing in $\Bcal_n$.  Hence, for every object $j \in \Ical^m_n$, we have $s^m_{j, c^m_{j,n}} = t_n$, and $\barz^m_{j, c^m_{j,n}} \overset{\Delta}{=} \z^m_{j}(t_n)$ is the factor state of the object at $t_n$. %Note that $s^m_{j, c^m_{j,n}-1}$ --- the preceding timestamp when the object showed up --- is not necessarily equal to $t_{n-1}$, since it might no appear in $\Bcal_{n-1}$ (see Fig. XX for an illustration). 
Upon receiving each $\Bcal_n$, we intend to approximate the running posterior of all the involved factor states and noise inverse variance $\tau$ with the following decoupled form, 
\begin{align}
p(\tau, \{\barz^m_{j, c^m_{j,n}}| j \in \Ical^m_n\}_{1 \le m \le M}|\Dcal_{t_n}) \approx q(\tau|\Dcal_n)\prod\nolimits_{m=1}^M \prod\nolimits_{j \in \Ical^m_n} q(\barz^m_{j, c^m_{j,n}}|\Dcal_{t_n}),  \label{eq:post-n}
\end{align}
where $q(\tau|\Dcal_{t_n}) = \text{Gam}(\tau | a_n, b_n)$, and $q(\barz^m_{j, c^m_{j,n}}|\Dcal_{t_n}) = \N(\barz^m_{j,c^m_{j, n}}|\widehat{\bmu}^m_{j,c^m_{j,n}}, \widehat{\V}^m_{j,c^m_{j,n}})$.
%\begin{align}
	%q(\barz^m_{j, c^m_{j,n}}|\Dcal_{t_n}) = \N(\barz^m_{j,c^m_{j, n}}|\bmu^m_{j,c^m_{j,n}}, \V^m_{j,c^m_{j,n}}).
%\end{align}
To this end, let us consider given the approximation at $t_n$, how to obtain the new approximation  at $t_{n+1}$ (\ie upon receiving $\Bcal_{n+1}$) in the same form of \eqref{eq:post-n}. To simplify the notation, let us define the preceding states of the involved factors by $\Theta_{n} = \{\barz^m_{j,c^m_{j,n}} | j \in \Ical^m_{n+1}\}_m$, and the current states by $\Theta_{n+1} = \{\barz^m_{j,c^m_{j,n+1}} | j \in \Ical^m_{n+1}\}_m$. 
First, due to the chain structure of the prior over each $\{\barz^m_{j,c^m_{j,k}} | k=0, 1, 2, \ldots\}$, we can see that  conditioned on $\{\Theta_n, \tau\}$,   the current states $\Theta_{n+1}$ and the new observations $\Bcal_{n+1}$ are independent  of $\Dcal_{t_n}$. This is because in the graphical model representation, $\{\Theta_n, \tau\}$ have blocked all the paths from the old observations $\Dcal_{t_n}$  to the new state and observations~\citep{Bishop07PRML}; see Fig. \ref{fig:graphical_model} for an illustration. Then, we can derive that 
\begin{align}
	&p(\Theta_{n+1}, \Theta_n, \tau |\Dcal_{t_{n+1}}) \propto p(\Theta_{n+1}, \Theta_{n}, \tau, \Bcal_{n+1}|\Dcal_{t_{n}}) =p(\Theta_n, \tau|\Dcal_{t_n}) p(\Theta_{n+1}, \Bcal_{n+1}|\Theta_{n}, \tau, \Dcal_{t_{n}}) \notag \\
	&=p(\Theta_n, \tau|\Dcal_{t_n}) p(\Theta_{n+1}|\Theta_{n}) p(\Bcal_{n+1}| \Theta_{n+1}, \tau),
\end{align}
where $p(\Theta_n, \tau|\Dcal_{t_n})$ is the running posterior at $t_n$, $$p(\Theta_{n+1}|\Theta_{n}) = \prod_{m=1}^M \prod_{j \in \Ical^m_{n+1}} p(\barz^m_{j,c^m_{j, n+1}}| \barz^m_{j,c^m_{j, n}}),$$ each $p(\barz^m_{j,c^m_{j, n+1}}| \barz^m_{j,c^m_{j, n}}) $ is a conditional Gaussian distribution  defined in \eqref{eq:state-space}, 
and  $p(\Bcal_{n+1}| \Theta_{n+1}, \tau)=  \prod_{(\bell, y) \in \Bcal_{n+1}} \N\left(y| \1^\top \left(\u^1_{\ell_1}(t_{n+1}) \circ \cdots \circ \u^M_{\ell_M}(t_{n+1})\right),  \tau^{-1}\right)$.
Since $p(\Theta_n, \tau|\Dcal_{t_n})$ takes the form of \eqref{eq:post-n}, we can analytically marginalize out each $\barz^m_{j, c^m_{j,n}} \in \Theta_{n}$, and obtain  
\begin{align}
	&p\left(\Theta_{n+1}, \tau|\Dcal_{t_{n+1}}\right) \propto \text{Gam}(\tau | a_n, b_n)   \prod\nolimits_{m=1}^M \prod\nolimits_{j \in \Ical^m_{n+1}}  \N(\barz^m_{j,c^m_{j, n+1}}|\widehat{\bmu}^{m}_{j,c^m_{j,n+1}}, \widehat{\V}^{m}_{j,c^m_{j,n+1}}) \label{eq:running-2}\\
	&  \cdot \prod\nolimits_{(\bell, y) \in \Bcal_{n+1}} \N\left(y| \1^\top \left(\u^1_{\ell_1}(t_{n+1}) \circ \cdots \circ \u^M_{\ell_M}(t_{n+1})\right), \notag \tau^{-1}\right). \notag 
\end{align}
If we view the R.H.S of \eqref{eq:running-2} as a joint distribution with $\Bcal_{n+1}$, then our task amounts to estimating the posterior distribution, \ie the L.H.S of \eqref{eq:running-2}. The product in the CP likelihood (and also Tucker likelihood) renders exact posterior computation infeasible, and we henceforth approximate  
\begin{align}
	&\N\left(y| \1^\top \left(\u^1_{\ell_1}(t_{n+1}) \circ \cdots \circ \u^M_{\ell_M}(t_{n+1})\right)\right) \appropto  \prod_{m=1}^M \N\left(\u^m_{\ell_m}(t_{n+1})|\bgamma^m_{\ell_m}, \bSigma^m_{\ell_m}\right) \text{Gam}(\tau|\alpha_{\bell}, \omega_{\bell}) \label{eq:ll-approx} 
\end{align}
where $\appropto$ means approximately proportional to. To optimize these approximation terms, we use the recent conditional Expectation propagation (CEP) framework~\citep{wang2019conditional} to develop an efficient inference algorithm. It uses conditional moment matching to update each approximation in parallel and conducts fixed point iterations, and hence can converge fast. We leave the details in the Appendix (Section \ref{sect:online-inference}). Once it is done, we substitute the approximation \eqref{eq:ll-approx} into \eqref{eq:running-2}. Then the R.H.S of \eqref{eq:running-2} becomes a product of  Gaussian and Gamma terms over each state and $\tau$. \cmt{Since both the Gaussian and Gamma distributions are in the exponential family, their product is still in the exponential family.} We can then immediately obtain a closed-form estimation in the form as \eqref{eq:post-n}. At the beginning, when estimating $p(\Theta_1, \tau|\Dcal_{t_1})$, since the preceding states $\Theta_0 = \emptyset$,  we have $a_n = \alpha_0$, $b_n = \alpha_1$, $\widehat{\bmu}^{m}_{j,c^m_{j,n+1}} = \0$, and $\widehat{\V}^{m}_{j,c^m_{j,n+1}} = \overline{\P}_\infty$ in \eqref{eq:running-2}, which is the prior of each $\barz^m_{j, c^m_{j,1}}$ and $\tau$ (see \eqref{eq:state-space}). 

In this way, we can continuously filter the incoming batches $\{\Bcal_1, \Bcal_2, \ldots\}$. As a result, for the factor state chain of every object $j$ in every mode $m$, along with each timestamp $s^m_{j, k}$, we can online estimate and track a running posterior approximation $\{q(\barz^m_{j, k} | \Dcal_{t_{s^m_{j,k}}})|k=1, 2, \ldots\}$, which is a Gaussian distribution. Hence, we can run the standard RTS smoother, to compute the full posterior of every factor state, with which we can compute the posterior of the trajectory at any time point $t$~\citep{bishop2006pattern}. Our method is summarized in Algorithm \ref{alg:alg}. 

\noindent\textbf{Algorithm Complexity.} The time complexity of our algorithm processing a batch $\Bcal_n$ is $\Ocal(|\Bcal_n| R^3)$ where $|\cdot |$ is the size. The time complexity of RTS smoother for a particular object $j$ in mode $m$ is $\Ocal(R^3 c^m_{j, N})$, where $N$ is the total number of timestamps. The space complexity of our algorithm is $\Ocal(\sum_{m=1}^M \sum_{j=1}^{d_m} c^m_{j, N} R^2)$, which is to track the running posterior of the factor state at each appearing timestamp for every object. Since $c^m_{j, N} \le N$, the complexity of our algorithm is at worst linear in $N$. 
\begin{algorithm}
	\caption{Streaming Factor Trajectory Learning (\ours)}
	\begin{algorithmic}[1]
		\STATE {\bfseries Input:} kernel hyper-parameters $a$, $\rho$, $\nu=p+\frac{1}{2}$ ($p \in \{0, 1, 2, \ldots\}$)
		\STATE $n \leftarrow 0$ 
		\WHILE {Receiving a new batch of entreis $\Bcal_{n+1}$}
		\IF{$n = 0$}
		\STATE Set $a_n = a_0$, $b_n = b_0$, $\widehat{\bmu}^{m}_{j,c^m_{j,n+1}} = \0$, and $\widehat{\V}^{m}_{j,c^m_{j,n+1}} = \overline{\P}_\infty$ in \eqref{eq:running-2}.
		\STATE Goto 9.
		\ENDIF
		\STATE  Retrieve the involved preceding factor states $\Theta_{n} = \{\barz^m_{j,c^m_{j,n}} | j \in \Ical^m_{n+1}\}_m$ and their running posterior, $p(\Theta_{n}, \tau|\Dcal_{t_{n}})\approx \text{Gam}(\tau|a_{n}, b_{n}) \prod_{m=1}^M \prod_{j \in \Ical^m_{n+1}} \N(\barz^m_{j, c^m_{j, n}} | \widehat{\bmu}^m_{j, c^m_{j, n}}, \widehat{\V}^m_{j, c^m_{j, n}}) $.
		%\begin{align}
		%	&p(\Theta_{n}, \tau|\Dcal_{t_{n}})\approx \text{Gam}(\tau|a_{n}, b_{n}) \notag \\
		%	&\cdot \prod_{m=1}^M \prod_{j \in \Ical^m_{n+1}} \N(\barz^m_{j, c^m_{j, n}} | \bmu^m_{j, c^m_{j, n}}, \V^m_{j, c^m_{j, n}}) \notag 
		%\end{align}
		%When $n=0$, $p(\Theta_n, \tau|\Dcal_{t_n})$ is the prior distribution. 
		\STATE  According to \eqref{eq:running-2} and \eqref{eq:ll-approx}, use conditional Expectation Propagation to calculate the running posterior of the current factor states, $p(\Theta_{n+1}, \tau|\Dcal_{t_{n+1}})\approx \text{Gam}(\tau|a_{n+1}, b_{n+1}) \prod_{m=1}^M \prod_{j \in \Ical^m_{n+1}} \N(\barz^m_{j, c^m_{j, n+1}} | \widehat{\bmu}^m_{j, c^m_{j, n+1}}, \widehat{\V}^m_{j, c^m_{j, n+1}})$.
		%\begin{align}
		%	&p(\Theta_{n+1}, \tau|\Dcal_{t_{n+1}})\approx \text{Gam}(\tau|a_{n+1}, b_{n+1}) \notag \\
		%	&\prod_{m=1}^M \prod_{j \in \Ical^m_{n+1}} \N(\barz^m_{j, c^m_{j, n+1}} | \bmu^m_{j, c^m_{j, n+1}}, \V^m_{j, c^m_{j, n+1}}) \notag 
		%\end{align}
		\IF {Needed}
		\STATE Run RTS smoothing on any factor  state chain $\{\barz^m_{j, k}|k=1, 2, \ldots\}$ of interest.
		\ENDIF
		\STATE $n \leftarrow n + 1$
		\ENDWHILE
		\STATE Run RTS smoothing for every factor state chain $\{\barz^m_{j,k}|k = 1, 2, \ldots \}$.
		\STATE {\bfseries Return:} $\{q(\barz^m_{j,k}|\Dcal)|k=1, 2, \ldots\}_{1 \le m \le M, 1 \le j \le d_m}$, $q(\tau|\Dcal)$, where $\Dcal$ is all the data received.
	\end{algorithmic}	\label{alg:alg}
\end{algorithm}

%% file: related.tex
%tensor factorizaton, temporal things
%EP & temporal GP applications 
\section{Related Work}
Many tensor decomposition methods have been developed, such as~\citep{YangDunson13Tensor,RaiDunson2014,zhe2015scalable,zhe2016distributed,zhe2016dintucker,tillinghast2020probabilistic,pan2020streaming,fang2021bayesian,fang2021streaming,tillinghast2021nonparametric,tillinghast2022nonparametric,fang2022bayesian,zhe2018stochastic,pan2020scalable,wang2020self,pan2021self,wang2022nonparametric}.
For temporal decomposition, most existing methods augment the tensor with a discrete time mode to  estimate additional factors for time steps, \eg ~\citep{xiong2010temporal,rogers2013multilinear,song2017multi,du2018probabilistic,ahn2021time}. The most recent works have conducted continuous-time decomposition. \citet{zhang2021dynamic} used polynomial splines to model  a time function as the CP coefficients. \citet{li2022decomposing} used neuralODE~\citep{chen2018neural} to model the entry value as a function of latent factors and time point. \citet{fang2022bayesian} performed continuous-time Tucker decomposition, and modeled the tensor-core as a time function. To our knowledge,  \citep{wang2022nonparametric} is the first work to estimate factor trajectories. It places a GP prior in the frequency domain, and samples the factor trajectories via inverse Fourier transform. It then uses another GP to sample the entry values. While successful,  this method cannot handle streaming data, and the black-box GP decomposition lacks interpretability. 

%streaming
Current Bayesian streaming tensor decomposition methods include~\citep{du2018probabilistic,fang2021bayesian,pan2020streaming,fang2021streaming}, which are based on streaming variational Bayes~\citep{broderick2013streaming} or assumed density filtering (ADF)~\citep{boyen1998tractable}.\cmt{ The works~\citep{pan2020streaming,fang2021bayesian} conduct nonlinear streaming decomposition based on GPs or Bayesian neural networks, and hence lack interpretability as well.} ADF can be viewed as an instance of Expectation Propagation (EP)~\citep{minka2001expectation}  for streaming data. EP approximates complex terms in the probability distribution with exponential-family members, and uses moment matching to iteratively update the approximations, which essentially is a fixed point iteration. To address the challenge of intractable moment matching, \citet{wang2019conditional} proposed conditional EP (CEP), which uses conditional moment matching and Taylor expansion to compute the moments for factorized approximations. The theoretical guarantees and error bound analysis for EP and ADF have been studied for a long time, such as \citep{boyen1998tractable,dehaene2015bounding,dehaene2018expectation}.
The most recent work \citep{fang2022bayesian} also uses SDEs to represent GPs and CEP framework for inference, but their GP prior is placed on the tensor-core, not for learning factor trajectories, and their method is only for static decomposition, and cannot handle streaming data.

%% file: exp-zhe.tex
\section{Experiment}
\subsection{Simulation Study}
We first conducted a simulation study, for which we simulated a two-mode tensor, with two nodes per mode. Each node is represented by a time-varying factor: $u^1_1(t) = -\sin^{3}(2\pi t)$,  $u^1_2(t) = \left(1-\sin^{3}(\frac{1}{2}\pi t)\right) \sin^{3}(3\pi t)$,  $u^2_1(t) = \sin(2\pi t)$, and $u^2_2(t) = -\cos^{3}(3\pi t) \sin(3\pi t)\sin(2\pi t)$.
%\begin{align}
%	&u^1_1(t) = -\sin^{3}(2\pi t), \; u^1_2(t) = \left(1-\sin^{3}(\frac{1}{2}\pi t)\right) \sin^{3}(3\pi t), \notag \\ 
%	&u^2_1(t) = \sin(2\pi t), \; u^2_2(t) = -\cos^{3}(3\pi t) \sin(3\pi t)\sin(2\pi t). \notag 
%\end{align}
Given these factors, an entry value at time $t$ is generated via $y_{(i, j)}(t) \sim \N\left( u^1_i(t) u^2_j(t), 0.05\right)$.
%\begin{align}
%y_{(i, j)}(t) = u^1_i(t) u^2_j(t) + \epsilon, \;\; \epsilon \sim \N(\cdot | 0, 0.05). \label{eq:sample-entry}
%\end{align}
\begin{figure*}
	\centering
	\setlength{\tabcolsep}{0pt}
	\begin{tabular}[c]{cccc}
		\begin{subfigure}[b]{0.25\textwidth}
			\centering
			\includegraphics[width=\linewidth]{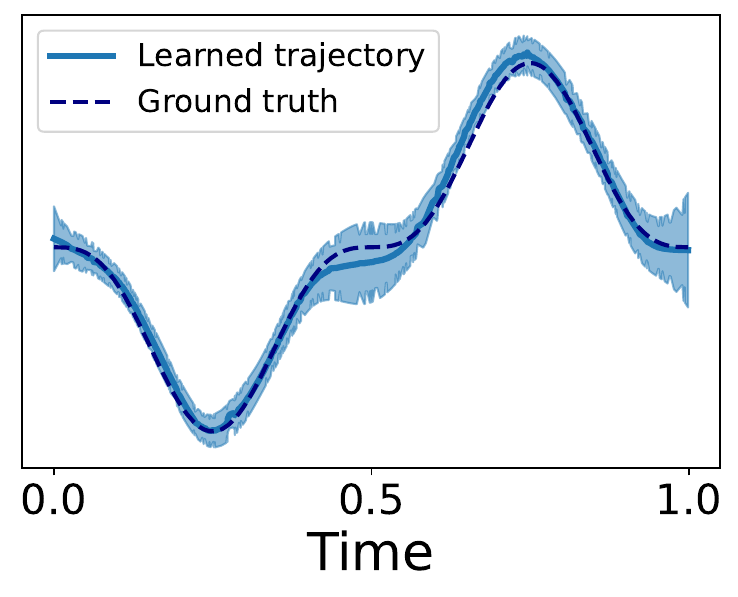}
			\caption{$u_1^1(t)$}
			\label{fig:simu-U11}
		\end{subfigure} &
		\begin{subfigure}[b]{0.25\textwidth}
			\centering
			\includegraphics[width=\linewidth]{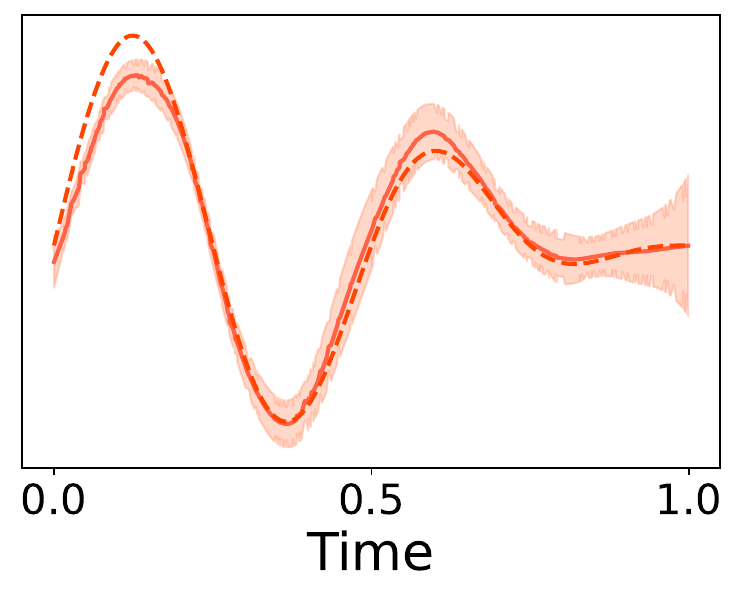}
			\caption{$u_2^1(t)$}
			\label{fig:simu-U12}
		\end{subfigure} &
		\begin{subfigure}[b]{0.25\textwidth}
			\centering
			\includegraphics[width=\linewidth]{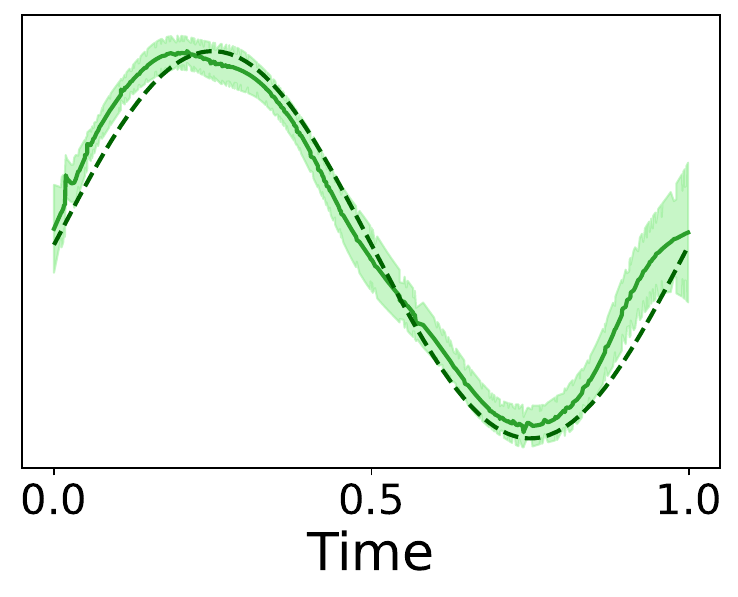}
			\caption{$u_1^2(t)$}
			\label{fig:simu-U21}
		\end{subfigure} &
		\begin{subfigure}[b]{0.25\textwidth}
			\centering
			\includegraphics[width=\linewidth]{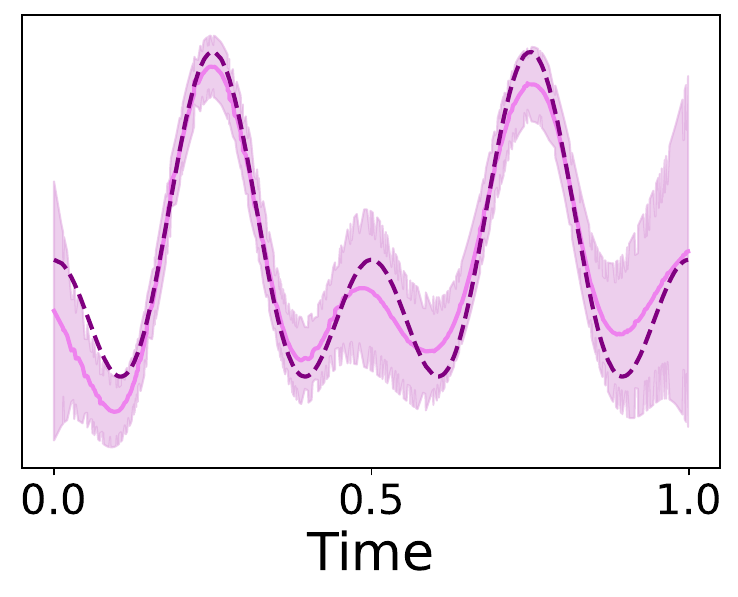}
			\caption{$u_2^2(t)$}
			\label{fig:simu-U22}
		\end{subfigure} 
	\end{tabular}
	\caption{\small The learned factor trajectories from the synthetic data. The shaded region indicates the posterior standard deviation. }\label{fig:simu-traj}
\end{figure*}
We randomly sampled $500$ (irregular) timestamps from $[0, 1]$. For each timestamp, we randomly picked two entries, and sampled their values accordingly. Overall, we sampled 1,000 observed tensor entry values. 

We implemented \ours with PyTorch~\citep{paszke2019pytorch}. We used $\nu = \frac{3}{2}$ and $a = \rho  = 0.3$ for the Mat\'ern kernel. We streamed the sampled entries according to their timestamps, and ran our streaming factor trajectory inference based on the CP form. The estimated trajectories  are shown in Fig. \ref{fig:simu-traj}.  As we can see, \ours recovers the ground-truth pretty accurately, showing that \ours has successfully captured the temporal evolution of the factor representation for every node. It is interesting to observe that when $t$ is around $0$, $0.5$ and $1$, the posterior standard deviation (the shaded region) increases significantly. This is reasonable: the ground-truth trajectories overlap at these time points, making it more difficult to differentiate/estimate their values at these time points. Accordingly, the uncertainty  of the estimation increases. In Section \ref{sect:simu-more} of Appendix, we further provide the root-mean-square error (RMSE) in recovering the four trajectories, and sensitivity analysis of the kernel parameters. 
%\textcolor{blue}{In Sec of Appendix, we also show the sensitivity of our method to the kernel parameters.} 

%\begin{figure}
\begin{table*}[]
	\centering
	\begin{small}
		\begin{tabular}{llcccc}
			\toprule
			& {RMSE}&  \textit{FitRecord} & \textit{ServerRoom} & \textit{BeijingAir-2} & \textit{BeijingAir-3}  \\ 
			\midrule
			\multirow{ 8}{*}{Static}&PTucker &    $ 0.656\pm0.147 $ & $ 0.458\pm0.039 $ & $ 0.401\pm0.01 $  & $ 0.535\pm0.062 $ \\
			&Tucker-ALS& $ 0.846\pm0.005 $ & $ 0.985\pm0.014 $ & $ 0.559\pm0.021 $ & $ 0.838\pm0.026 $ \\
			&CP-ALS  &       $ 0.882\pm0.017 $ & $ 0.994\pm0.015 $ & $ 0.801\pm0.082 $ & $ 0.875\pm0.028 $ \\
			&CT-CP       &               $ 0.664\pm0.007 $ & $ 0.384\pm0.009 $ & $ 0.64\pm0.007 $  & $ 0.815\pm0.018 $ \\
			&CT-GP          &            $ 0.604\pm0.004 $ & $ 0.223\pm0.035 $ & $ 0.759\pm0.02 $  & $ 0.892\pm0.026 $ \\
			&BCTT        &           $ 0.518\pm0.007 $ & $ 0.185\pm0.013 $ & $ 0.396\pm0.022 $ & $ 0.801\pm0.02 $  \\
			&NONFAT     &            $ 0.503\pm0.002 $ & $ \mathbf{0.117\pm0.006 }$ & $ 0.395\pm0.007 $ & $ 0.882\pm0.014 $ \\
			&THIS-ODE    &  $ 0.526\pm0.004 $         & $ 0.132\pm0.003 $ & $ 0.54\pm0.014 $         & $ 0.877\pm0.026 $ \\ 
			\hline
			\multirow{5}{*}{Stream}&POST        &            $ 0.696\pm0.019 $ & $ 0.64\pm0.028 $  & $ 0.516\pm0.028 $ & $ 0.658\pm0.103 $ \\
			&ADF-CP   &            $ 0.648\pm0.008 $ & $ 0.654\pm0.008 $ & $ 0.548\pm0.015 $ & $ 0.551\pm0.043 $ \\
			&BASS-Tucker        &            $ 0.976\pm0.024 $ & $ 1.000\pm0.016 $     & $ 1.049\pm0.037 $ & $ 0.991\pm0.039 $ \\
			&SFTL-CP     &  $ \mathbf{0.424\pm0.014} $ & $ 0.161\pm0.014 $ & $ \mathbf{0.248\pm0.012} $ & $ 0.473\pm0.013 $ \\
			&SFTL-Tucker &  $ 0.430\pm0.010 $   & $ 0.331\pm0.056 $ & $ 0.303\pm0.041 $ & $ \mathbf{0.439\pm0.019} $ \\ 
			\midrule
			& {MAE}    &           \\ 
			\midrule
			\multirow{ 8}{*}{Static}&PTucker &            $ 0.369\pm0.009 $ & $ 0.259\pm0.008 $ & $ 0.26\pm0.006 $  & $ 0.263\pm0.02 $  \\
			&Tucker-ALS &  $ 0.615\pm0.006 $ & $ 0.739\pm0.008 $ & $ 0.388\pm0.008 $ & $ 0.631\pm0.017 $ \\
			&CP-ALS  &            $ 0.642\pm0.012 $ & $ 0.746\pm0.009 $ & $ 0.586\pm0.056 $ & $ 0.655\pm0.018 $ \\
			&CT-CP       &            $ 0.46\pm0.004 $  & $ 0.269\pm0.003 $ & $ 0.489\pm0.006 $ & $ 0.626\pm0.01 $  \\
			&CT-GP          & $ 0.414\pm0.001 $ & $ 0.165\pm0.034 $ & $ 0.55\pm0.012 $  & $ 0.626\pm0.011 $ \\
			&BCTT        &            $ 0.355\pm0.005 $ & $ 0.141\pm0.011 $ & $ 0.254\pm0.007 $ & $ 0.578\pm0.009 $ \\
			&NONFAT     &            $ 0.341\pm0.001 $ & $ \mathbf{0.071\pm0.004} $ & $ 0.256\pm0.004 $ & $ 0.626\pm0.007 $ \\
			&THIS-ODE    & $ 0.363\pm0.004 $         & $ 0.083\pm0.002 $ & $ 0.345\pm0.004 $         & $ 0.605\pm0.013 $ \\
			\hline
			\multirow{5}{*}{Stream}&POST        &            $ 0.478\pm0.014 $ & $ 0.476\pm0.023 $ & $ 0.352\pm0.022 $ & $ 0.486\pm0.095 $ \\
			&ADF-CP              & $ 0.449\pm0.006 $ & $ 0.496\pm0.007 $ & $ 0.385\pm0.012 $ & $ 0.409\pm0.029 $ \\
			&BASS        &           $ 0.772\pm0.031 $ & $ 0.749\pm0.01 $  & $ 0.934\pm0.037 $ & $ 0.731\pm0.02 $  \\
			&SFTL-CP   &   $ \mathbf{0.242\pm0.006} $ & $ 0.108\pm0.008 $ & $\mathbf{ 0.15\pm0.003} $  & $ 0.318\pm0.008 $ \\
			&SFTL-Tucker &         $ 0.246\pm0.001 $ & $ 0.216\pm0.034 $ & $ 0.185\pm0.029 $ & $ \mathbf{0.278\pm0.011} $ \\ 
			\bottomrule
		\end{tabular}
	\end{small}
	\caption{\small Final prediction error \cmt{of the static and streaming decomposition methods,} with $R=5$. The results were averaged from five runs.}
	\label{table:Rank5res}
	\vspace{-0.3in}
\end{table*}

\begin{figure*}[!ht]
	\centering
	\setlength\tabcolsep{0pt}
	\begin{tabular}[c]{cccc}
		%\multicolumn{4}{c}{\includegraphics[width=0.718\textwidth]{./figs_new/legend2.pdf}}\\
		\begin{subfigure}[t]{0.25\textwidth}
			\centering
			\includegraphics[width=\textwidth]{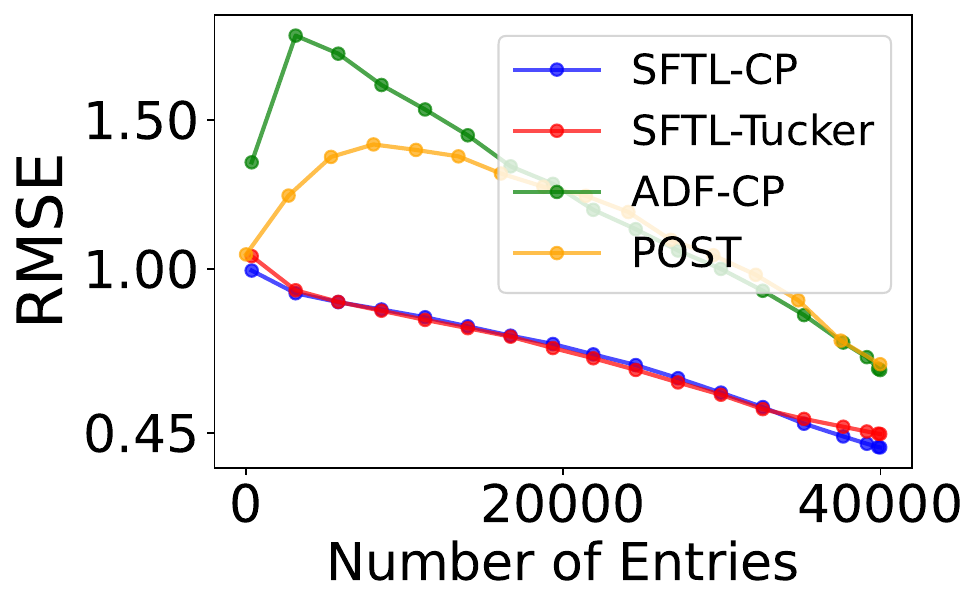}
			\caption{\textit{FitRecord} }
		\end{subfigure} 
		&
		\begin{subfigure}[t]{0.25\textwidth}
			\centering
			\includegraphics[width=\textwidth]{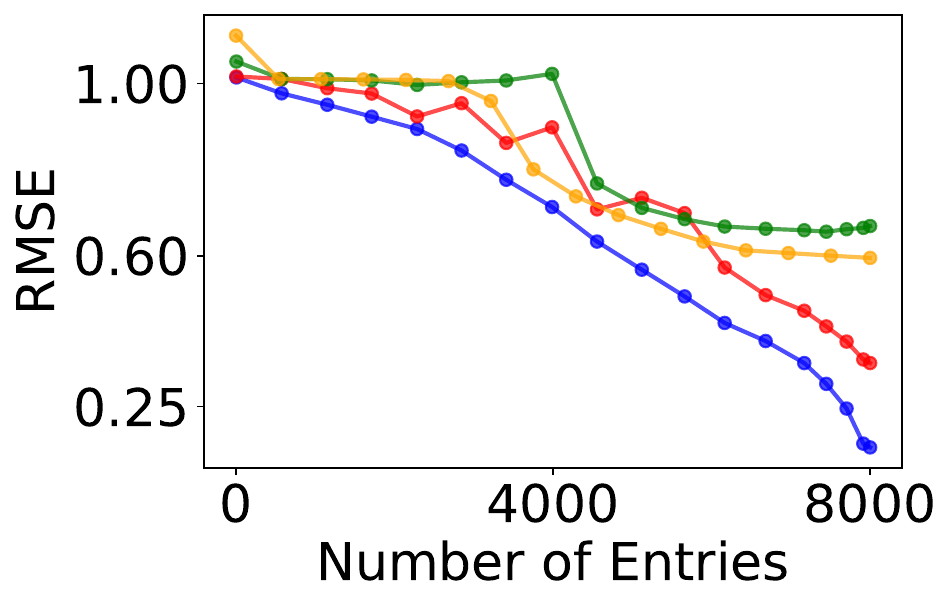}
			\caption{\textit{ServerRoom}}
		\end{subfigure} 
		&
		\begin{subfigure}[t]{0.25\textwidth}
			\centering
			\includegraphics[width=\textwidth]{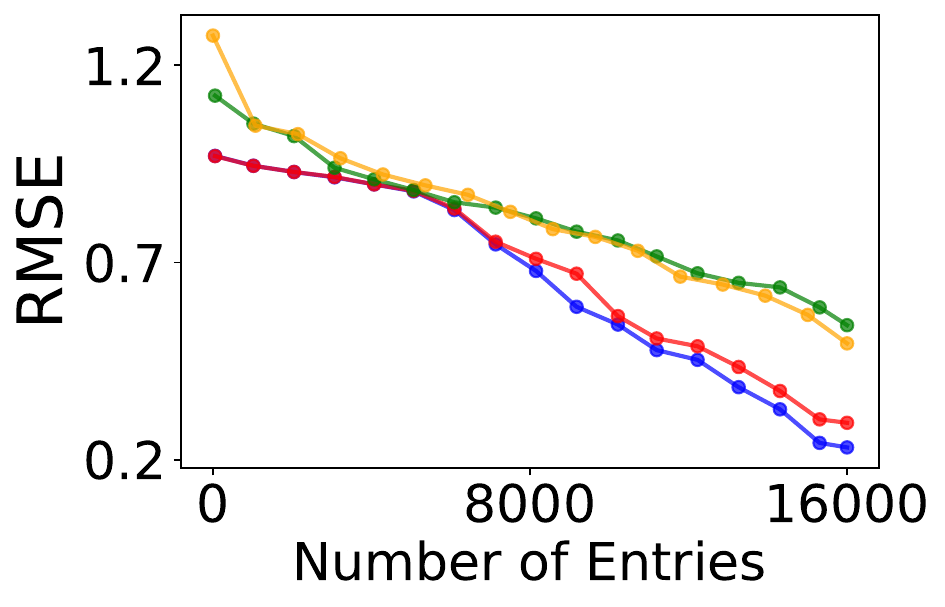}
			\caption{\textit{BeijingAir-2} }
		\end{subfigure}
		&
		\begin{subfigure}[t]{0.25\textwidth}
			\centering
			\includegraphics[width=\textwidth]{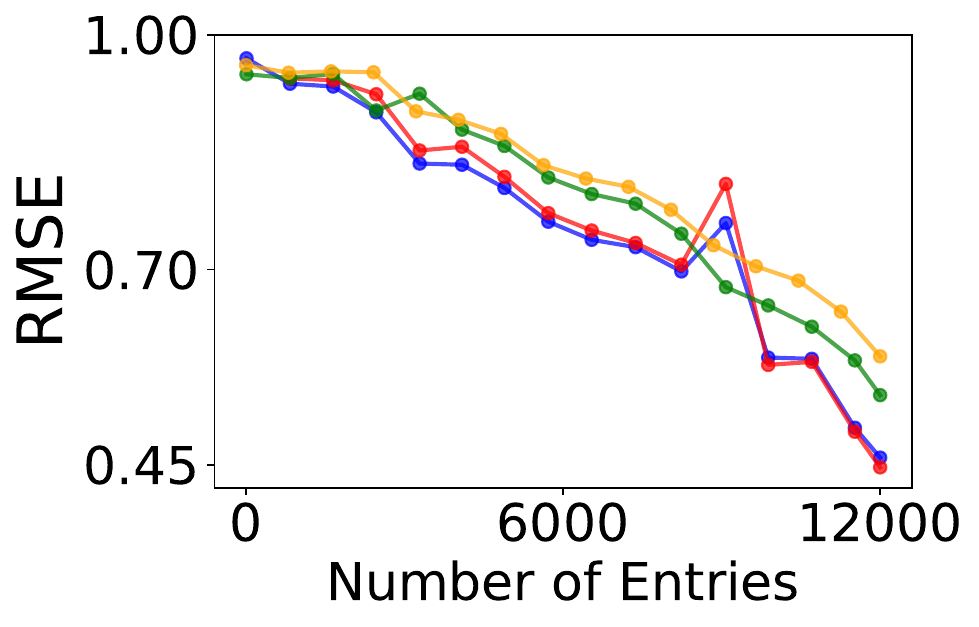}
			\caption{\textit{BeijingAir-3} }
		\end{subfigure}
	\end{tabular}	
	\caption{\small Online prediction error with the number of processed entries ($R=5$).} 	
	\label{fig:running-performance-r5}
	\vspace{-0.2in}
\end{figure*}

\subsection{Real-World Applications}
Next, we examined \ours in four real-world datasets: \textit{FitRecord}, \textit{ServerRoom}, \textit{BeijingAir-2}, and \textit{BeijingAir-3}. We tested 11 competing approaches.  We compared with state-of-the-art streaming tensor decomposition methods based on the CP or Tucker model, including (1) {POST} ~\citep{du2018probabilistic},  (2) {BASS-Tucker} ~\citep{fang2021bayesian} and (3) ADF-CP~\citep{wang2019conditional}, the state-of-the-art static decomposition algorithms, including (4) P-Tucker~\citep{oh2018scalable},  (5) CP-ALS and (6) Tucker-ALS~\citep{bader2008efficient}.   For those methods, we augment the tensor  with a time mode, and convert the ordered, unique timestamps into increasing time steps. We also compared with the most recent continuous-time decomposition methods.  (7)  {CT-CP}~\citep{zhang2021dynamic},  (8) {CT-GP}, (9) {BCTT} ~\citep{fang2022bayesian},  (10) {THIS-ODE}~\citep{li2022decomposing}, and (11)  {NONFAT}~\citep{wang2022nonparametric}, nonparametric factor trajectory learning, the only existing work that also estimates factor trajectories for temporal tensor decomposition. %It uses a bi-level GP to estimate the trajectories in the frequency domain and applies inverse Fourier transform to return to the time domain. 
Note that the methods 4-11 cannot handle data streams. They have to iteratively access the data to update the model parameters and factor estimates. The details about the competing methods and datasets are provided in Appendix (Section \ref{sect:expr-details}).

%{\bf Settings.}
For all the competing methods, we used the publicly released  implementations of the original authors. The hyper-parameter setting and turning follows the original papers.  For \ours, we chose $\nu$ from $\{\frac{1}{2}, \frac{3}{2}\}$,  $a$ from [0.5, 1] and $\rho$ from [0.1, 0.5]. For our online filtering, the maximum number of CEP iterations was set to $50$ and the tolerance level to $10^{-4}$.  For numerical stability, we re-scaled the timestamps to $[0, 1]$. We examined the number of factors (or factor trajectories) $R \in \{2, 3, 5, 7\}$. 

%\subsection{Final Prediction Accuracy}
\noindent\textbf{Final Prediction Accuracy.} We first examined the final prediction accuracy with our learned factor trajectories. To this end, we followed ~\citep{xu2012infinite,kang2012gigatensor}, and randomly sampled $80\%$ observed entry values and their timestamps for streaming inference and then tested the prediction error on the remaining entries. We also compared with the static decomposition methods, which need to repeatedly access the training entries. We repeated the experiment five times, and computed the average root mean-square-error (RMSE), average mean-absolute-error (MAE), and their standard deviations. We ran our method based on both the CP and Tucker forms, denoted by \ours-CP and \ours-Tucker, respectively. 
\begin{figure*}
	%	\vspace{-0.2in}
	\centering
	\setlength{\tabcolsep}{0pt}
	\begin{tabular}[c]{ccc}
		\begin{subfigure}[b]{0.3\textwidth}
			\centering
			\includegraphics[width=\linewidth]{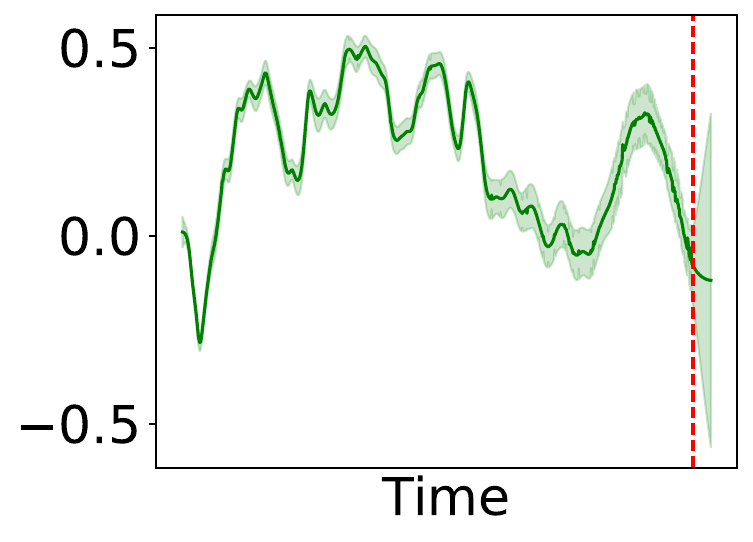}
			\caption{$u_{1,1}^{1}(t)$}
			\label{fig:traj-U1-1}
		\end{subfigure} &
		\begin{subfigure}[b]{0.3\textwidth}
			\centering
			\includegraphics[width=\linewidth]{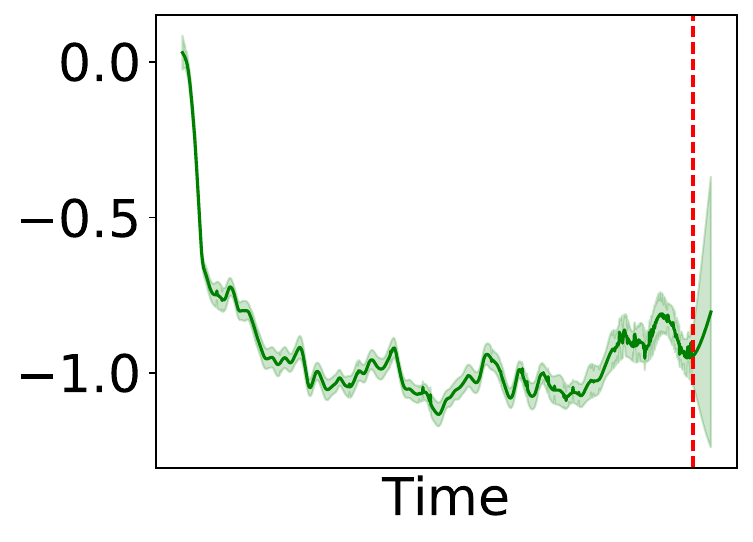}
			\caption{$u_{1,2}^{1}(t)$}
			\label{fig:traj-U1-2}
		\end{subfigure} &
		\begin{subfigure}[b]{0.3\textwidth}
			\centering
			\includegraphics[width=\linewidth]{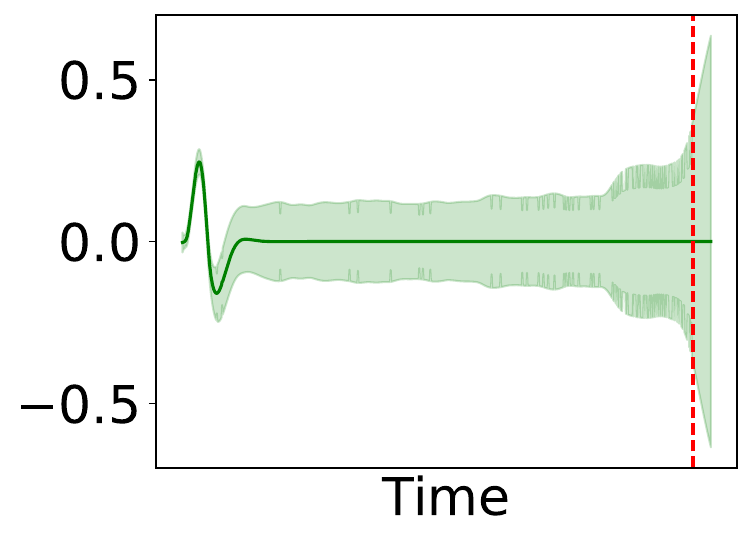}
			\caption{$u_{1,3}^{1}(t)$}
			\label{fig:traj-U1-3}
		\end{subfigure} \\
		\begin{subfigure}[b]{0.3\textwidth}
			\centering
			\includegraphics[width=\linewidth]{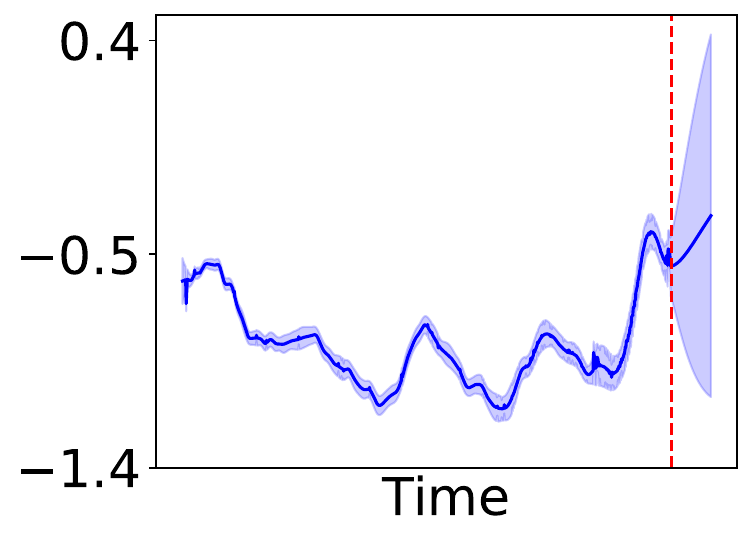}
			\caption{$u_{2,1}^{2}(t)$}
			\label{fig:traj-U2-1}
		\end{subfigure} &
		\begin{subfigure}[b]{0.3\textwidth}
			\centering
			\includegraphics[width=\linewidth]{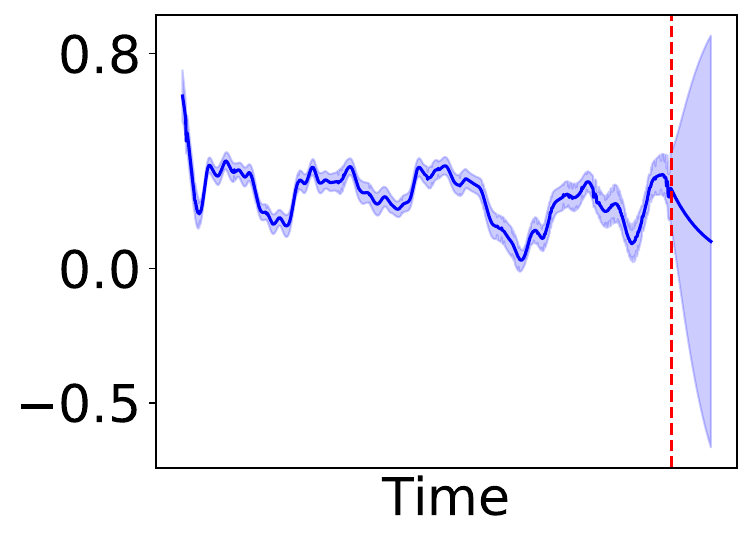}
			\caption{$u_{2,2}^{2}(t)$}
			\label{fig:traj-U2-2}
		\end{subfigure} &
		\begin{subfigure}[b]{0.3\textwidth}
			\centering
			\includegraphics[width=\linewidth]{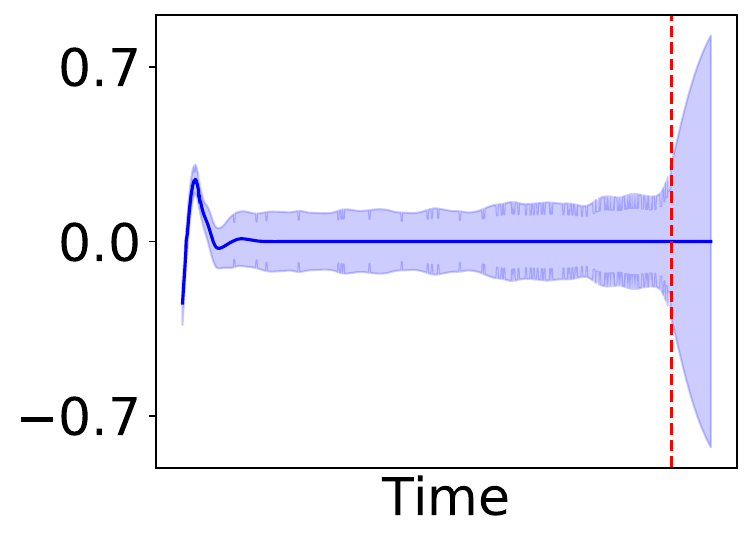}
			\caption{$u_{2,3}^{2}(t)$}
			\label{fig:traj-U2-3}
		\end{subfigure} \\
		\begin{subfigure}[b]{0.3\textwidth}
			\centering
			\includegraphics[width=\linewidth]{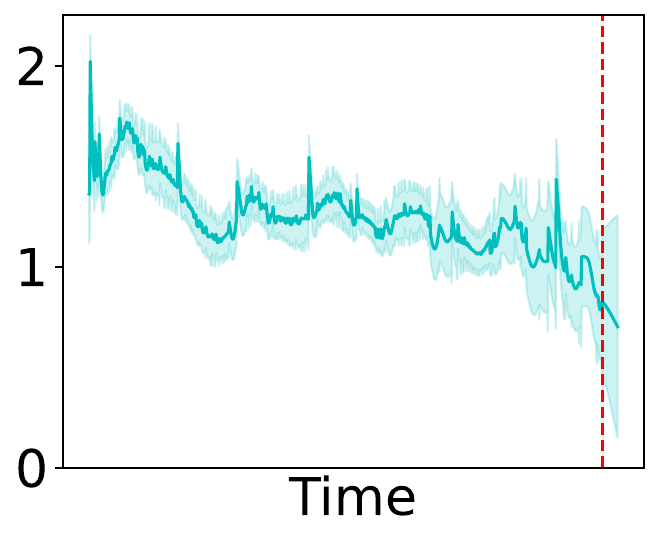}
			\caption{$u_{3,1}^{3}(t)$}
			\label{fig:traj-U3-1}
		\end{subfigure} &
		\begin{subfigure}[b]{0.3\textwidth}
			\centering
			\includegraphics[width=\linewidth]{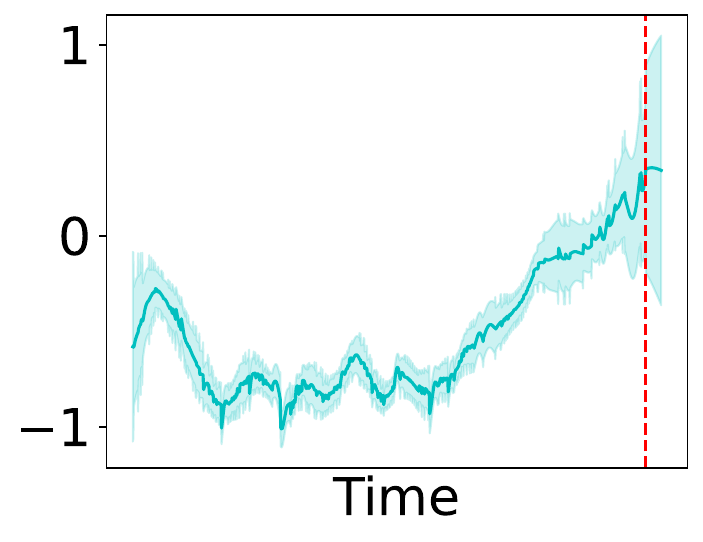}
			\caption{$u_{3,2}^{3}(t)$}
			\label{fig:traj-U3-2}
		\end{subfigure} &
		\begin{subfigure}[b]{0.3\textwidth}
			\centering
			\includegraphics[width=\linewidth]{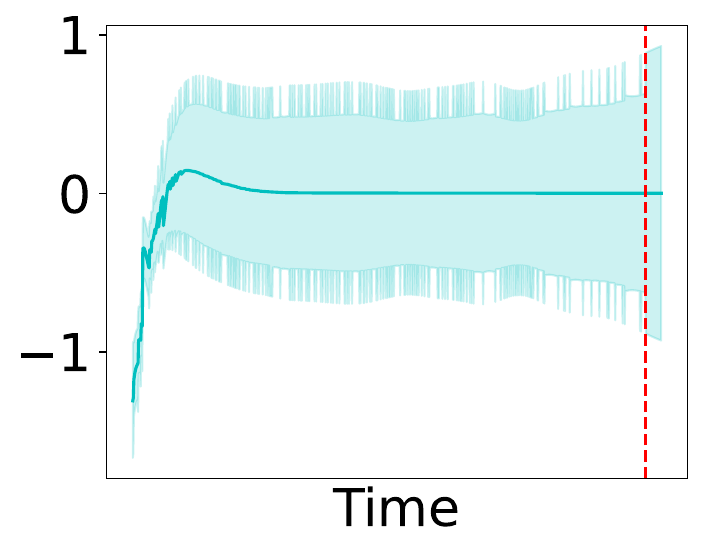}
			\caption{$u_{3,3}^{3}(t)$}
			\label{fig:traj-U3-3}
		\end{subfigure} 
	\end{tabular}
	\caption{\small The learned factor trajectories of object 1, 2, 3 in mode 1, 2, 3, respectively, from  \textit{ServerRoom} data. The shaded region shows one standard deviation and the dashed line indicates the biggest timestamp in the data.}
	\label{fig:traj}
\end{figure*}
We report the results for $R=5$ in Table \ref{table:Rank5res}. Due to the space limit, we leave the other results in the Appendix (Table \ref{table:Rank2res},\ref{table:Rank3res}, and \ref{table:Rank7res}). As we can see, \ours outperforms all the streaming approaches by a large margin. \ours even obtains significantly better prediction accuracy than all the static decomposition approaches, except that on \textit{Server Room}, \ours is second to THIS-ODE and NONFAT. Note that \ours only went through the training entries for once. Although NONFAT  can also estimate the factor trajectories, it uses GPs to perform black-box nonlinear decomposition and hence loses  the interpretability. Note that NONFAT in most case also  outperforms the other static decomposition methods that only estimate time-invariant factors. The superior performance of \ours and NONFAT shows the importance of capturing factor evolution. 

%\subsection{Online Predictive Performance}
\noindent \textbf{Online Predictive Performance.} Next, we evaluated the online predictive performance of \ours. Whenever a batch of entries at a new timestamp has been processed, we examined the prediction accuracy on the test set, with our current estimate of the factor trajectories.   We repeated the evaluation for five times, and examine how the average prediction error varies along with the number of processed entries. We show the results for $R=5$ in Fig. \ref{fig:running-performance-r5}, and the others in Appendix (Fig. \ref{fig:running-performance-r2}, \ref{fig:running-performance-r3}, and \ref{fig:running-performance-r7}). It is clear that \ours in most cases outperforms the competing streaming decomposition algorithms by a large margin throughout the course of running. Note that the online behavior of BASS-Tucker was quite unstable and so we excluded it in the figures. 
%Although the other methods also exhibit decreasing prediction errors with more and more entries were processed, their running performance is constantly worse  than \ours. 
It confirms the advantage of our streaming trajectory learning approach ---  even in the streaming scenario, incrementally capturing the time-variation of the factors can perform better than updating fixed, static factors. %\textcolor{blue}{we show the computational efficiency in Sec of Appendix}. 
To confirm the advantage of \ours in computational efficiency, we report the running time in Section \ref{sect:running-time} of Appendix. 

%\subsection{Investigation of Learning Results }

\noindent\textbf{Investigation of Learning Results.} Finally, we investigated our learned factor trajectories. We set $R = 3$ and ran \ours-CP on \textit{ServerRoom}. In Fig. \ref{fig:traj}, we visualize the trajectory for the 1st air conditioning mode, the 2rd power usage level,  and the 3rd location. The shaded region indicates the standard deviation, and the red dashed line the biggest timestamp in the data. First, we can see that the posterior variance of all the trajectories grow quickly  when moving to the right of the red line. This is reasonable because it is getting far away from the training region. Second, for each object, the first and second trajectories (1st and 2nd column in Fig. \ref{fig:traj}) exhibit quite different time-varying patterns, \eg the local periodicity in $u^1_{1, 2}(t)$ and $u^2_{2,2}(t)$, the gradual decreasing and increasing trends in $u^3_{3,1}(t)$ and $u^3_{3,2}(t)$, respectively, which imply different inner properties of the object. Third, it is particularly interesting to see that the third trajectory for all the objects appear to be close to zero, with relatively large posterior variance all the time. This might imply that two factor trajectories have been sufficient to represent each object.  Requesting for a third trajectory is redundant.  More important, our model is empirically able to detect such redundancy and returns a zero-valued trajectory. 

%% file: conclusion.tex
\vspace{-0.05in}
\section{Conclusion}
\vspace{-0.05in}
We have presented \ours, a probabilistic temporal tensor decomposition approach.  \ours can efficiently handle streaming data, and estimate time-varying factor representations. On four real-world applications, \ours achieves superior online and final prediction accuracy.

\section*{Acknowledgments}
This work has been supported by MURI AFOSR
grant FA9550-20-1-0358 and NSF CAREER Award
IIS-2046295.

%% file: supp-zhe.tex
\section*{Appendix}
\section{Spectral Analysis and LTI-SDE}\label{appendix:sect:A}
We consider the Mat\'ern kernel family, 
\begin{align}
	\kappa_\nu\left(t, t'\right)&=a \frac{\left(\frac{\sqrt{2 \nu}}{\rho} \Delta \right)^{\nu}}{\Gamma(\nu) 2^{\nu-1}}
	K_{\nu}\left(\frac{\sqrt{2 \nu}}{\rho} \Delta\right), \label{Matern_kernel}
\end{align}
where $\Delta = |t - t'|$, $\Gamma(\cdot)$ is the Gamma function, $a>0$ and $\rho>0$ are the amplitude and length-scale parameters, respectively, $K_{\nu}$ is the modified Bessel function of the second kind, $\nu>0$ controls the smoothness.\cmt{ It means that if $f(t) \sim \gp(0, \kappa_\nu(t, t'))$, $f(t)$ is $\lceil \nu \rceil-1$ times differentiable in the mean-square sense []. }
Since $\kappa_\nu$ is a stationary kernel, \ie $\kappa_\nu(t, t') = \kappa_\nu(t - t')$, according to the  Wiener-Khinchin theorem~\citep{chatfield2003analysis}, if $$f(t) \sim \gp(0, \kappa_\nu(t, t')),$$ the energy spectrum density of $f(t)$ can be obtained by the Fourier transform of $\kappa_\nu(\Delta)$, 
\begin{align}
	S(\omega ) = a \frac{2\sqrt{\pi}\Gamma(\frac{1}{2} + \nu)}{\Gamma(\nu)} \alpha^{2\nu}\left(\alpha^2 + \omega^2\right)^{-\left(\nu + \frac{1}{2}\right)},
\end{align}
where $\omega$ is the frequency, and $\alpha = \frac{\sqrt{2\nu}}{\rho}$. We consider the commonly used choice  $\nu = p + \frac{1}{2}$ where $p \in \{0, 1, 2, \ldots\}$. Then we can observe that 
\begin{align}
	S(\omega) = \frac{\sigma^2}{(\alpha^2 + \omega^2)^{p+1}} = \frac{\sigma^2}{(\alpha + i \omega)^{p+1}(\alpha - i \omega)^{p+1}}, \label{eq:spec-density}
\end{align}
where $\sigma^2 = a\frac{ 2\sqrt{\pi}\Gamma(p+1)}{\Gamma(p+\frac{1}{2})} \alpha^{2p + 1}$, and $i$ indicates an imaginary number. We expand the polynomial 
\begin{align}
	(\alpha + i \omega)^{p+1} = \sum\nolimits_{k=0}^p c_k (i\omega)^k + (i\omega)^{p+1}, \label{eq:poly}
\end{align}
where $\{c_k|0 \le k \le p\}$ are the coefficients.  From \eqref{eq:spec-density} and \eqref{eq:poly}, we can construct an equivalent system to generate the signal $f(t)$. That is,  in the frequency domain, the system output's Fourier transform   $\widehat{f}(\omega)$ is given by 
\begin{align}
	\sum\nolimits_{k=1}^p c_k (i\omega)^k \widehat{f}(\omega) + (i\omega)^{p+1} \widehat{f}(\omega) = \widehat{\beta}(\omega), \label{eq:fourier}
\end{align}
where $\widehat{\beta}$ is the Fourier transform of a white noise process $\beta(t)$ with spectral density (or diffusion) $\sigma^2$. The reason is that by construction, $\widehat{f}(\omega) = \frac{\widehat{\beta}(\omega)}{(\alpha + i \omega)^{p+1}}$, which gives exactly the same spectral density as in \eqref{eq:spec-density},  $S(\omega) = |\widehat{f}(\omega)|^2$.\cmt{ $= \widehat{f}(\omega)\widehat{f}^\dagger(\omega)$, where $\dagger$ is the complex conjugate.}  We then conduct inverse Fourier transform on both sides of \eqref{eq:fourier} to obtain the representation in the time domain, 
\begin{align}
	\sum\nolimits_{k=1}^p c_k \frac{\d^k f}{\d t^k} + \frac{\d^{p+1}f}{\d t^{p+1}} = \beta(t), \label{eq:sde}
\end{align}
which is an SDE. Note that $\beta(t)$ has the density $\sigma^2$. We can further construct a new state $\z = (f, f^{(1)}, \ldots, f^{(p)})^\top$ (where each $f^{(k)} \overset{\Delta}{=} \d^k f/\d t^k$) and convert \eqref{eq:sde} into a linear time-invariant (LTI) SDE, 
\begin{align}
	\frac{\d \z}{\d t} = \A \z + \boldeta \cdot \beta(t),
\end{align}
where 
\begin{align}
	\A =\left(\begin{array}{cccc}
		0 & 1 & & \\
		& \ddots  & \ddots & \\
		&   & 0 & 1 \\
		-c_0 & \ldots & -c_{p-1} & -c_{p}
	\end{array}\right), \quad \boldeta =\left(\begin{array}{c}
		0 \\
		\vdots\\
		0 \\
		1\\
	\end{array}\right). \notag 
\end{align}
For a concrete example, if we take $p=1$ (and so $\nu = \frac{3}{2}$), then $\A = [0, 1; -\alpha^2, -2 \alpha]$, $\boldeta = [0; 1]$, and $\sigma^2 = 4 a \alpha^3$.

The LTI-SDE is particularly useful  in that its finite set of states follow a Gauss-Markov chain, namely the state-space prior.  Specifically, given arbitrary $t_1 < \ldots < t_L$, we have 
\begin{align}
	p(\z(t_1), \ldots, \z(t_L)) = p(\z(t_1)) \prod\nolimits_{k=1}^{L-1} p(\z(t_{k+1})| \z(t_{k})), \notag 
\end{align}
where $p(\z(t_1)) = \N(\z(t_1)|\0, \P_\infty)$, $p(\z(t_{k+1})| \z(t_{k})) = \N(\z(t_{k+1}) |\F_{k} \z(t_{k}), \Q_k)$, 
%\begin{align}
%	p(\z(t_1)) &= \N(\z(t_1)|\0, \P_\infty), \notag \\
%	p(\z(t_{k+1})| \z(t_{k})) &= \N(\z(t_{k+1}) |\F_{k} \z(t_{k}), \Q_k)
%\end{align}
$\P_\infty$ is the stationary covariance matrix computed by solving the matrix Riccati equation~\citep{lancaster1995algebraic}\cmt{ (\eg $\P_\infty = [a, 0; 0, a\alpha^2]$ when $n=1$)}, $\F_n = \exp(\Delta_k \cdot \A )$ where $\Delta_k = t_{k+1} - t_k$, and $\Q_k = \P_\infty - \A_k \P_\infty \A_k^\top$.  Therefore, we  do not  need the full covariance matrix as in the standard GP prior, and the computation is much more efficient. The chain structure is also convenient to handle streaming data as explained in the main paper. 

Note that for other type of kernel functions, such as the square exponential (SE) kernel, we can approximate the inverse spectral density $1/S(\omega)$ with a polynomial of $\omega^2$ with negative roots, and follow the same way to construct an LTI-SDE (approximation) and state-space prior. 

\section{RTS Smoother} \label{sect:rts}
Consider a standard state-space model with state $\x_n$ and observation $\y_n$ at each time step $n$. The prior distribution is a Gauss-Markov chain, $$p(\x_{n+1}|\x_n) = \N(\x_{n+1}|\A_n\x_n, \Q_n),$$  $$p(\x_0) = \N(\x_0|\m_0, \P_0).$$ Suppose we have a Gaussian observation likelihood, $$p(\y_n|\x_n) = \N(\y_n|\H_n \x_n, \W_n).$$ Then upon receiving each $\y_n$, we can use Kalman filtering to obtain the exact running posterior, $$p(\x_n | \y_{1:n})= \N(\x_n | \m_k, \P_k),$$ which is a Gaussian.  After all the data has been processed --- suppose it ends after step $N$ --- we can use Rauch–Tung–Striebel (RTS) smoother~\citep{sarkka2013bayesian} to efficiently compute the full posterior of each state from backward, which does not need to re-access any data:   $p(\x_n|\y_{1:N}) = \N(\x_n | \m_n^s, \P_n^s)$, where 
\begin{align}
	\m^-_{n+1} &= \A_n \m_n, \;\; \P^-_{n+1} = \A_n \P_n \A_n^\top + \Q_n , \notag \\
	\G_n &= \P_n \A_n^\top [\P^-_{n+1}]^{-1}, \notag \\
	\m^s_n &= \m_n + \G_n\left(\m^s_{n+1} - \m^-_{n+1}\right), \notag \\
	\P^s_n &= \P_n + \G_n[\P^s_{n+1} - \P^-_{n+1}]\G_n^\top. \label{eq:rts}
\end{align}
As we can see, the computation only needs the running posterior $p(\x_n|\y_{1:n})=\N(\cdot|\m_n, \P_n)$ and the full posterior of the next state $p(\x_{n+1}|\y_{1:N}) = \N(\cdot | \m_{n+1}, \P_{n+1})$. It does not need to revisit previous observations $\y_{1:N}$

\section{Details about Online Trajectory Inference}\label{sect:online-inference}
In this section, we provide the details about how to update the running posterior according to \eqref{eq:running-2} and \eqref{eq:ll-approx} (in the main paper) with the conditional EP (CEP) framework~\citep{wang2019conditional}.
\subsection{EP and CEP framework}
We first give a brief  introduction to the EP and CEP framework. Consider a general probabilistic model with latent parameters $\btheta$. Given the observed data $\Dcal = \{\y_1, \ldots, \y_N\}$, the joint probability distribution is 
\begin{align}
	p(\btheta, \Dcal) = p(\btheta) \prod_{n=1}^N p(\y_n|\btheta).
\end{align}
Our goal is to compute the posterior $p(\btheta|\Dcal)$. However, it is usually infeasible to compute the exact marginal distribution $p(\Dcal)$, because of the complexity of the likelihood and/or prior. EP therefore seeks to approximate each term in the joint probability by an exponential-family term, 
\begin{align}
	p(y_n|\btheta) \approx   c_n f_n(\btheta), \;\;\; p(\btheta) \approx  c_0	f_0(\btheta),
\end{align}
where $c_n$ and $c_0$ are constants to ensure the normalization consistency (they will get canceled in the inference, so we do not need to calculate them), and 
$$f_n(\btheta) \propto \exp(\blambda_n^\top \bphi(\btheta)), (0 \le n \le N)$$
where $\blambda_n$ is the natural parameter and $\bphi(\btheta)$ is sufficient statistics. For example, if we choose a Gaussian term, $f_n = \N(\btheta| \bmu_n, \bSigma_n)$, then the sufficient statistics is $\bphi(\btheta) = \{\btheta, \btheta\btheta^\top\}$. The moment is the expectation of the sufficient statistics.%, \ie the mean and covariance 

We therefore approximate the joint probability with 
\begin{align}
	&p(\btheta, \Dcal) = p(\btheta) \prod_{n=1}^N p(\y_n|\btheta) \approx f_0(\btheta) \prod_{n=1}^N f_n(\btheta) \cdot \text{const}. \label{eq: approx-joint}
\end{align}
Because the exponential family is closed under product operations, we can immediately obtain a closed-form approximate posterior $q(\btheta) \approx p(\btheta|\Dcal)$ by merging the approximation terms in the R.H.S of \eqref{eq: approx-joint}, which is still a distribution in the exponential family.  

Then the task amounts to optimizing those approximation terms $\{f_n(\btheta) | 0 \le n \le N\}$. EP repeatedly conducts four steps to optimize each $f_n$. 
\begin{itemize}
	\item \textbf{Step 1.} We obtain the calibrated distribution that integrates the context information of $f_n$, 
	\begin{align}
		q^{\backslash n}(\btheta) \propto \frac{q(\btheta)}{f_n(\btheta)}, \notag 
	\end{align}
	where $q(\btheta)$ is the current posterior approximation. 
	\item \textbf{Step 2.} We construct a tilted distribution to combine the true likelihood,  
	\begin{align}
		 \tp(\btheta) \propto q^{\bkh n} (\btheta) \cdot p(\y_n | \btheta). \notag 
	\end{align}
	Note that if $ n = 0$, we have $\tp(\btheta) \propto q^{\bkh n} (\btheta) \cdot p( \btheta)$.
	\item \textbf{Step 3.} We project the tilted distribution back to the exponential family, $$q^*(\btheta) = \argmin_q \;\; \kl( \tp \| q)$$ where $q$ belongs to the exponential family. This can be done by moment matching, 
	\begin{align}
		\EE_{q^*}[\bphi(\btheta)] = \EE_\tp[\bphi(\btheta)]. \label{eq:mm}
	\end{align}
	That is, we compute the expected moment under $\tp$, with which to obtain the parameters of $q^*$.  For example, if $q^*(\btheta)$ is a Gaussian distribution, then we need to compute $\EE_\tp[\btheta]$ and $\EE_\tp[\btheta\btheta^\top]$, with which to obtain the mean and covariance for $q^*(\btheta)$. Hence we obtain $q^*(\btheta) = \N(\btheta| \EE_\tp[\btheta],\EE_\tp[\btheta\btheta^\top] -  \EE_\tp[\btheta] \EE_\tp[\btheta]^\top )$
	\item \textbf{Step 4.} We update the approximation term by 
	\begin{align}
		f_n(\btheta) \approx \frac{q^*(\btheta)}{q^\bkh(\btheta)}.
	\end{align}
\end{itemize}
In practice, EP often updates all the $f_n$'s in parallel, and uses damping to avoid divergence. It iteratively runs the four steps until convergence. In essence, this is a fixed point iteration to optimize a free energy function (a mini-max problem)~\citep{minka2001expectation}.

The critical step in EP is the moment matching \eqref{eq:mm}. However, in many cases, it is analytically intractable to compute the moment under the tilted distribution $\tp$, due to the complexity of the likelihood. To address this problem, CEP considers the commonly used case that each $f_n$ has a factorized structure, 
\begin{align}
	f_n(\btheta) = \prod_m f_{nm}(\btheta_m),
\end{align}
where each $f_{nm}$ is also in the exponential family, and $\{\btheta_m\}$ are mutually disjoint. Then at the moment matching step, we need to compute the moment of each $\btheta_m$ under $\tp$, \ie $\EE_\tp[\bphi(\btheta_m)]$. The first key idea of CEP is to use the nested structure, 
\begin{align}
	\EE_\tp[\bphi(\btheta_m)] = \EE_{\tp(\btheta_{\bkh m})} \EE_{\tp(\btheta_m|\btheta_{\bkh m})} [\bphi(\btheta_m)],
\end{align}
where $\btheta_{\bkh m} = \btheta \bkh \btheta_m$. Therefore, we can first compute the inner expectation, \ie conditional moment, 
	\begin{align}
		\EE_{\tp(\btheta_m|\btheta_{\bkh m})} [\bphi(\btheta_m)] = \g(\btheta_{\bkh m}),
	\end{align}
and then seek for computing the outer expectation, $\EE_{\tp(\btheta_{\bkh m})} [\g(\btheta_{\bkh m})]$. The inner expectation is often easy to compute (\eg with our CP/Tucker likelihood). When $f_n$ is factorized individually over each element of $\btheta$, this can always be efficiently and accurately calculated by quadrature. However, the outer expectation is still difficult to obtain  because $\tp(\btheta_{\bkh m})$ is intractable. The second key idea of CEP is that since the moment matching is also between $q(\btheta_{\bkh m})$ and $\tp(\btheta_{\bkh m})$, we can use the current marginal posterior to approximate the marginal titled distribution and then compute the outer expectation, 
\begin{align}
	\EE_{\tp(\btheta_{\bkh m})} [\g(\btheta_{\bkh m})] \approx \EE_{q(\btheta_{\bkh m})}[\g(\btheta_{\bkh m})].
\end{align}
If it is still analytically intractable, we can use the delta method~\citep{oehlert1992note} to approximate the expectation.  That is, we use a Taylor expansion of $\g(\cdot)$ at the mean of $\btheta_{\bkh m}$. Take the first-order expansion as an example, 
\begin{align}
	\g(\btheta_{\bkh m})  \approx \g\left(\EE_{q(\btheta_{\bkh m})}[\btheta_{\bkh m}]\right) + \J\left(\btheta_{\bkh m} - \EE_{q(\btheta_{\bkh m})}[\btheta_{\bkh m}]\right)  \notag 
\end{align} 
where $\J$ is the Jacobian of $\g$ at $\EE_{q(\btheta_{\bkh m})}[\btheta_{\bkh m}]$.  Then we take the expectation on the Taylor approximation instead, 
\begin{align}
	\EE_{q(\btheta_{\bkh m})}\left[\g(\btheta_{\bkh m})\right] \approx \g\left(\EE_{q(\btheta_{\bkh m})}[\btheta_{\bkh m}]\right). \label{eq:delta}
\end{align}

The above computation is very convenient to implement. Once we obtain the conditional moment $\g(\btheta_{\bkh m})$, we simply replace the $\btheta_{\bkh m}$ by its expectation under current posterior approximation $q$, \ie $\EE_{q(\btheta_{\bkh m})}[\btheta_{\bkh m}]$, to obtain the matched moment $\g(\EE_{q(\btheta_{\bkh m})}[\btheta_{\bkh m}])$, with which to construct $q^*$ in Step 3 of EP (see \eqref{eq:mm}). The remaining steps are the same. 

\subsection{Running Posterior Update}
Now we use the CEP framework to update the running posterior $p(\Theta_{n+1}, \tau |\Dcal_{t_{n+1}})$  in \eqref{eq:running-2} via the approximation \eqref{eq:ll-approx}. To simplify the notation, let us define $\v^m_{l_m} \overset{\Delta}{=} \u^m_{\ell_m}(t_{n+1})$, and hence  for each $(\bell, y) \in \Bcal_{n+1}$, we approximate 
\begin{align}
	&\N\left(y|\1^\top \left (\v^1_{\ell_1} \circ \ldots \circ \v^M_{\ell_M}\right), \tau^{-1}\right) \approx  \prod_{m=1}^M \N(\v^m_{\ell_m}|\bgamma^m_{\ell_m}, \bSigma^m_{\ell_m}) \text{Gam}(\tau|\alpha_\bell, \omega_\bell). \label{eq:ll-approx-good}
\end{align}
If we substitute \eqref{eq:ll-approx} into \eqref{eq:running-2}, we can immediately obtain a Gaussian posterior approximation of each $\v^m_{\ell_m}$ and a Gamma posterior approximation of the noise inverse variance $\tau$. Then dividing the current posterior approximation with the R.H.S of \eqref{eq:ll-approx-good}, we can obtain the calibrated distribution, 
\begin{align}
	q^{\bkh \bell}(\v^m_{\ell_m}) &= \N(\v^m_{\ell_m}| \bbeta^m_{\ell_m}, \bOmega^m_{\ell_m}), \notag \\
	q^{\bkh \bell}(\tau) &= \text{Gam}(\tau | \alpha^{\bkh \bell}, \omega^{\bkh \bell}),
\end{align}
where $1 \le m \le M$.  Next, we construct a tilted distribution, 
\begin{align}
	&\tp(\v^1_{\ell_1}, \ldots, \v^M_{\ell_M}, \tau) \propto q^{\bkh \bell}(\tau)  \cdot \prod_{m=1}^M q^{\bkh \bell}(\v^m_{\ell_m}) \cdot \N\left(y|\1^\top \left (\v^1_{\ell_1} \circ \ldots \circ \v^M_{\ell_M}\right), \tau^{-1}\right).
\end{align}
To update each $\N(\v^m_{\ell_m}|\bgamma^m_{\ell_m}, \bSigma^m_{\ell_m})$ in \eqref{eq:ll-approx-good}, we first look into the conditional tilted distribution, 
\begin{align}
	&\tp(\v^m_{\ell_{m}}|\Vcal^{\bkh m}_{\bell}, \tau) \propto \N(\v^m_{\bell_m}|\bbeta^m_{\ell_m}, \bOmega^m_{\ell_m}) \cdot \N\left(y|\left(\v^m_{\ell_m}\right)^\top \v^{\bkh m}_\bell, \tau^{-1}\right)
\end{align}
where $\Vcal^{\bkh m}_{\bell}$ is $\{\v^j_{\ell_j}| 1 \le j \le M, j \neq m\}$, and $$\v^{\bkh m}_{\bell} = \v^1_{\ell_1} \circ \ldots \circ \v^{m-1}_{\ell_{m-1}} \circ \v^{m+1}_{\ell_{m+1}} \circ \ldots \circ \v^M_{\ell_M}.$$
The conditional tilted distribution is obviously Gaussian, and the conditional moment is straightforward to obtain, 
\begin{align}
	&\S(\v^m_{\ell_m}| \Vcal^{\bkh m}_\bell, \tau) = \left[{\bOmega^m_{\ell_m}}^{-1} + \tau \v^{\bkh m}_\bell \left(\v^{\bkh m}_\bell\right)^\top\right]^{-1}, \label{eq:cond-mm-1} \\
	& \EE[\v^m_{\ell_m}|\Vcal^{\bkh m}_\bell, \tau] = \S(\v^m_{\ell_m}| \Vcal^{\bkh m}_\bell, \tau)\cdot \left({\bOmega^m_{\bell_m}}^{-1}\bbeta^m_{\ell_m} + \tau y \v^{\bkh m}_\bell\right), \label{eq:cond-mm-2}
\end{align}
where $\S$ denotes the conditional covariance. Next, according to \eqref{eq:delta}, we simply replace $\tau$, $\v^{\bkh m}_\bell $, and $\v^{\bkh m}_\bell \left(\v^{\bkh m}_\bell\right)^\top$ by their expectation under the current posterior $q$ in \eqref{eq:cond-mm-1} and \eqref{eq:cond-mm-2}, to obtain the moments, \ie the mean and covariance matrix, with which we can construct $q^*$ in Step 3 of the EP framework. The computation of $\EE_q[\tau]$ is straightforward,  and 
\begin{align}
	&\EE_q[\v^{\bkh m}_\bell] = \EE_q[\v^1_{\ell_1} ] \circ \ldots \circ \EE_q[\v^{m-1}_{\ell_{m-1}} ] \circ  \EE_q[\v^{m+1}_{\ell_{m+1}} ]\circ  \ldots \circ  \EE_q[\v^{M}_{\ell_{M}} ],  \notag \\
	& \EE_q[\v^{\bkh m}_\bell\left(\v^{\bkh m}_\bell\right)^\top] =\EE_q[\v^1_{\ell_1} \left(\v^1_{\ell_1}\right)^\top]  \circ \ldots \circ \EE_q[\v^{m-1}_{\ell_{m-1}} \left(\v^{m-1}_{\ell_{m-1}}\right)^\top] \notag \\
	&\circ \EE_q[\v^{m+1}_{\ell_{m+1}} \left(\v^{m+1}_{\ell_{m+1}}\right)^\top] \circ \ldots \circ \EE_q[\v^M_{\ell_M} \left(\v^M_{\ell_M}\right)^\top]. \notag 
\end{align}

Similarly, to update $\text{Gam}(\alpha_{\bell}, \omega_\bell)$ in \eqref{eq:ll-approx-good}, we first observe that the conditional titled distribution is also a Gamma distribution, 
\begin{align}
	&\tp(\tau|\Vcal_\bell) \propto \text{Gam}(\tau|\talpha, \tomega) \propto \text{Gam}(\tau|\alpha^{\bkh \bell}, \omega^{\bkh \bell}) \N(y|\1^\top \v_\bell, \tau^{-1}),
\end{align} 
where $\v_\bell = \v^1_{\ell_1} \circ \ldots \circ \v^M_{\ell_M}$, and 
\begin{align}
	\talpha &= \alpha^{\bkh \bell} + \frac{1}{2}, \notag \\
	\tomega &= \omega^{\bkh \bell} + \frac{1}{2} y^2 + \frac{1}{2} \1^\top \v_\bell \v_\bell^\top \1 - y \1^\top \v. \label{eq:cond-moment-tau}
\end{align}
Since the conditional moments (the expectation of $\tau$ and $\log \tau$) are functions  of $\alpha$ and $\omega$, when using the delta method to approximate the expected conditional moment, it is equivalent to approximating the expectation of $\talpha$ and $\tomega$ first, and then use the expected $\talpha$ and $\tomega$ to recover the moments. As a result, we can simply replace $\v_\bell$ and $\v_\bell \v_\bell^\top$  in \eqref{eq:cond-moment-tau} by their expectation under the current posterior, and we obtain the approximation of $\EE_q[\talpha]$ and $\EE_q[\tomega]$. With these approximated expectation, we then construct $q^*(\tau) = \text{Gam}(\tau|\EE_q[\alpha], \EE_q[\omega])$ at Step 3 in EP. The remaining steps are straightforward. The running posterior  update with the Tucker form likelihood follows a similar way. 

\section{More Results on Simulation Study}\label{sect:simu-more}

\subsection{Accuracy of Trajectory Recovery}
We provide the quantitative result in recovering the factor trajectories. Note that there is only one competing method, NONFAT, which can also estimate factor trajectories.  We therefore ran our method and NONFAT on the synthetic dataset. We then randomly sampled 500 time points in the domain and evaluate the RMSE of the learned factor trajectories for each method. 
As shown in Table \ref{table:Trajectories-recover}, the RMSE of NONFAT on recovering $u^1_1(t)$ and $u^2_1(t)$ is close to \ours, showing NONFAT achieved the same (or very close) quality in recovering these two trajectories. However, on $u^1_2(t)$ and $u^2_2(t)$, the RMSE of NONFAT is much larger, showing that NONFAT have failed to capture the other two trajectories. By contrast, \ours consistently well recovered them. 

\begin{table*}[]
	\centering
	\begin{small}
		\begin{tabular}{lcccc}
			\toprule
			{  } & $u_1^1(t)$ & $u_2^1(t)$ & $u_1^2(t)$ & $u_2^2(t)$ \\
			\hline
			{  SFTL}                      & {  0.073}         & {  0.082}         & {  0.103}         & {  0.054}         \\
			{  NONFAT}                  & {  0.085}         & {  0.442}         & {  0.096}         & {  0.443}         \\
			\bottomrule
		\end{tabular}
	\end{small}
	%\vspace{-0.1in}
	\caption{\small RMSE in recovering trajectories on the simulation data.}
	\label{table:Trajectories-recover}
\end{table*}

\subsection{Sensitive Analysis on Kernel Parameters}
To examine the sensitivity to the kernel parameters, we used the synthetic dataset, and randomly sampled 100 entries and new timestamps for evaluation. We then examined the length-scale $\rho$ and amplitude $a$, for two commonly-used Mat\'ern kernels: Mat\'ern-1/2 and Mat\'ern-3/2. The study was performed on \ours based on both the CP and Tucker forms. The results are reported in Table \ref{table:Sensitive-analysis}. Overall, the predictive performance of \ours is less sensitive to the amplitude parameter $a$ than to the length-scale parameter $\rho$.  But when we use Mat\'ern-1/2, the performance of both \ours-CP and \ours-Tucker is quite stable to the length-scale parameter $\rho$. When we use Mat\'ern-3/2, the choice of the length-scale is critical. 
\begin{table*}[]
	\centering
	\begin{small}
		\begin{subtable}{\linewidth}
			\centering
		\begin{tabular}{lcccccc}
			\toprule
			 &  $\rho$ & {0.1}& {0.3} & {0.5} & {0.7} & {0.9}                 \\
			\hline 
			\multirow{2}{*}{Mat\'ern-1/2} & {  SFTL-CP}     & {  0.091} & {  0.064} & {  0.059} & {  0.056} & {  0.057} \\
			           & {  \ours-Tucker} & {  0.060} & {  0.055} & {  0.056} & {  0.056} & {  0.057} \\
			 \hline
			\multirow{2}{*}{Mat\'ern-3/2} & {  SFTL-CP}     & {  0.062} & {  0.061} & {  0.074} & {  0.093} & {  0.112} \\
			           & {  \ours-Tucker} & {  0.061} & {  0.059} & {  0.078} & {  0.101} & {  0.129} \\
			\bottomrule
		\end{tabular}
	\caption{\small Prediction RMSE with $a = 0.3$ and varying $\rho$.}
	\end{subtable}
	\begin{subtable}{\linewidth}
		\centering
		\begin{small}
		\begin{tabular}{lcccccc}
			\toprule
			 & $a$ & {0.1}& {0.3} & {0.5} & {0.7} & {0.9}                 \\
			 \hline
			\multirow{2}{*}{Mat\'ern-1/2}         & {   SFTL-CP}            & {   0.056}        & {   0.064}        & {   0.057}        & {   0.059}        & {   0.063}        \\
			                   & {   SFTL-Tucker}        & {   0.065}        & {   0.055}        & {   0.054}        & {   0.055}        & {   0.055}        \\
			\multirow{2}{*}{Mat\'ern-3/2}         & {   SFTL-CP}            & {   0.072}        & {   0.061}        & {   0.063}        & {   0.060}        & {   0.059}        \\
			                   & {   SFTL-Tucker}        & {   0.098}        & {   0.059}        & {   0.064}        & {   0.062}        & {   0.061}       \\
			\bottomrule
		\end{tabular}
	\end{small}
\caption{\small  Prediction RMSE with $\rho = 0.3$ and varying $a$.}
	\end{subtable}
	\end{small}
	%\vspace{-0.1in}
	\caption{\small Sensitive analysis of amplitude $a$ and length-scale $\rho$ on synthetic data.}
	\label{table:Sensitive-analysis}
	%\vspace{-0.1in}
\end{table*}

\section{Real-World Dataset Information and Competing Methods}\label{sect:expr-details}
We tested all the methods in the following four real-world datasets. 
\begin{itemize}
	\item  \textit{FitRecord}\footnote{\url{https://sites.google.com/eng.ucsd.edu/fitrec-project/home}}, workout logs of EndoMondo users' health status in outdoor exercises. We extracted a three-mode tensor among 500 users, 20 sports types, and 50 altitudes. The entry values are heart rates. There are $50$K observed entry values along with the timestamps.
	\item \textit{ServerRoom}\footnote{\url{https://zenodo.org/record/3610078\#\%23.Y8SYt3bMJGi}},   temperature logs of Poznan Supercomputing and Networking Center. We extracted a three-mode tensor between 3 air conditioning modes (24$^\circ$, 27$^\circ$ and 30$^\circ$), 3 power usage levels (50\%, 75\%, 100\%) and 34 locations. We collected 10K entry values and their timestamps.
	\item \textit{BeijingAir-2}\footnote{\url{https://archive.ics.uci.edu/ml/datasets/Beijing+Multi-Site+Air-Quality+Data}}, air pollution measurement in Beijing from year 2014 to 2017. We extracted a two-mode tensor (monitoring site, pollutant), of size $12 \times 6$, and collected  $20$K observed entry values (concentration) and their timestamps.
	\item \textit{BeijingAir-3}, extracted from the same data source as \textit{BeijingAir-2}, a three-mode tensor among 12 monitoring sites, 12 wind speeds and 6 wind directions. The entry value is the PM2.5 concentration. There are $15$K observed entry values at different timestamps.
\end{itemize}

We first compared with the following state-of-the-art streaming tensor decomposition methods based on the CP or Tucker model. (1) {POST} ~\citep{du2018probabilistic}, probabilistic streaming CP decomposition via mean-field streaming variational Bayes~\citep{broderick2013streaming} (2) {BASS-Tucker} ~\citep{fang2021bayesian} Bayesian streaming Tucker decomposition, which online estimates a sparse tensor-core via a spike-and-slab prior to enhance the interpretability. We also implemented (3) ADF-CP, streaming CP decomposition by combining  the assumed density filtering and conditional moment matching~\citep{wang2019conditional}.  %These methods online estimate a static factor representation for the objects appearing in the observed tenor entries, and they do not consider timestamps of the entries. To incorporate the time information for a fair comparison, we 

Next, we tested the state-of-the-art static decomposition algorithms, which have to go through the data many times. (4) P-Tucker~\citep{oh2018scalable}, an efficient Tucker decomposition algorithm that performs parallel row-wise updates. (5) CP-ALS and (6) Tucker-ALS~\citep{bader2008efficient},  CP/Tucker decomposition via alternating least square (ALS) updates. The methods  (1-6) are not specifically designed for temporal decomposition and cannot utilize the timestamps of the observed entries. In order to incorporate the time information for a fair comparison, we augment the tensor  with a time mode, and convert the ordered, unique timestamps into increasing time steps. 

We then compared with the most recent continuous-time temporal decomposition methods. Note that none of these methods can handle data streams. They have to iteratively access the data to update the model parameters and factor estimates. (7)  {CT-CP}~\citep{zhang2021dynamic}, continuous-time CP decomposition, which uses polynomial splines to model a time-varying coefficient $\blambda$ for each latent factor, (8) {CT-GP}, continuous-time GP decomposition, which extends~\citep{zhe2016dintucker} to use GPs to learn the tensor entry value as a function of the latent factors and time $y_\bell(t) = g(\u^1_{\ell_1}, \ldots, \u^K_{\ell_K}, t) \sim \gp(0,\kappa(\cdot, \cdot))$, (9) {BCTT} ~\citep{fang2022bayesian}, Bayesian continuous-time Tucker decomposition, which estimates the tensor-core as a time-varying function, (10) {THIS-ODE}~\citep{li2022decomposing}, which uses a neural ODE~\citep{chen2018neural} to model the entry  value as a function of the latent factors and time, $\frac{\d y_\bell(t)}{\d t} = \text{NN}(\u^1_{\ell_1}, \ldots, \u^K_{\ell_K}, t)$ where NN is short for neural networks. (11)  {NONFAT}~\citep{wang2022nonparametric}, nonparametric factor trajectory learning, the only existing work that also estimates factor trajectories for temporal tensor decomposition. It uses a bi-level GP to estimate the trajectories in the frequency domain and applies inverse Fourier transform to return to the time domain.

\section{More Results about Prediction Accuracy}

\begin{figure*}[!ht]
	\centering
	\setlength\tabcolsep{0.01pt}
	\begin{tabular}[c]{cccc}
		%\multicolumn{4}{c}{\includegraphics[width=0.718\textwidth]{./figs_new/legend2.pdf}}\\
		\begin{subfigure}[t]{0.24\textwidth}
			\centering
			\includegraphics[width=\textwidth]{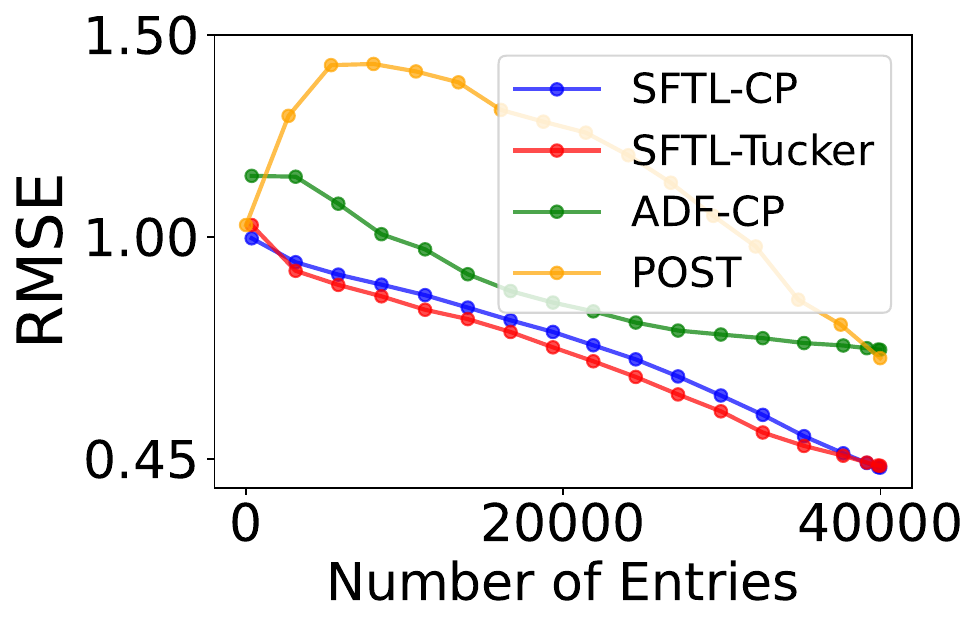}
			\caption{\textit{FitRecord}}
		\end{subfigure} 
		&
		\begin{subfigure}[t]{0.24\textwidth}
			\centering
			\includegraphics[width=\textwidth]{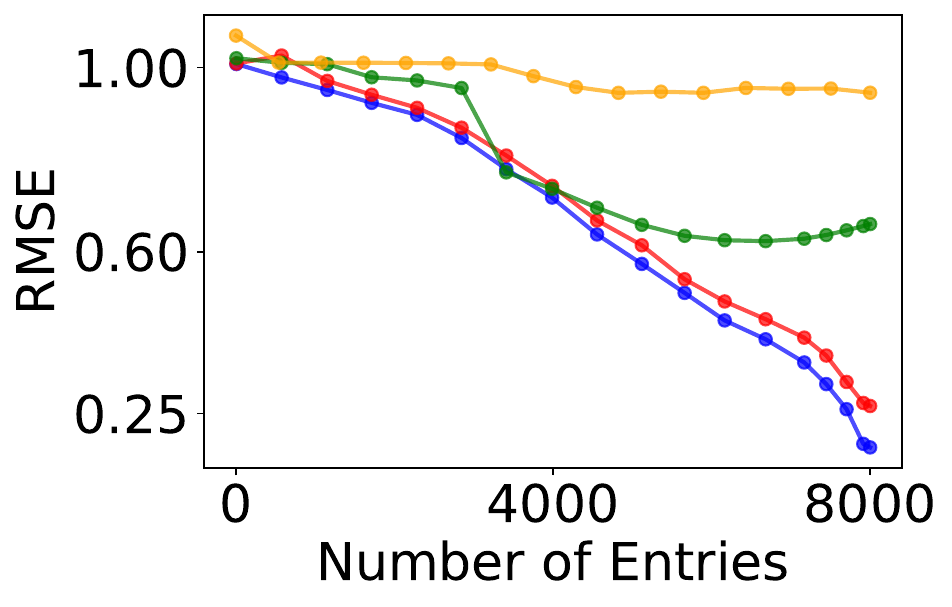}
			\caption{\textit{ServerRoom} }
		\end{subfigure} 
		&
		\begin{subfigure}[t]{0.24\textwidth}
			\centering
			\includegraphics[width=\textwidth]{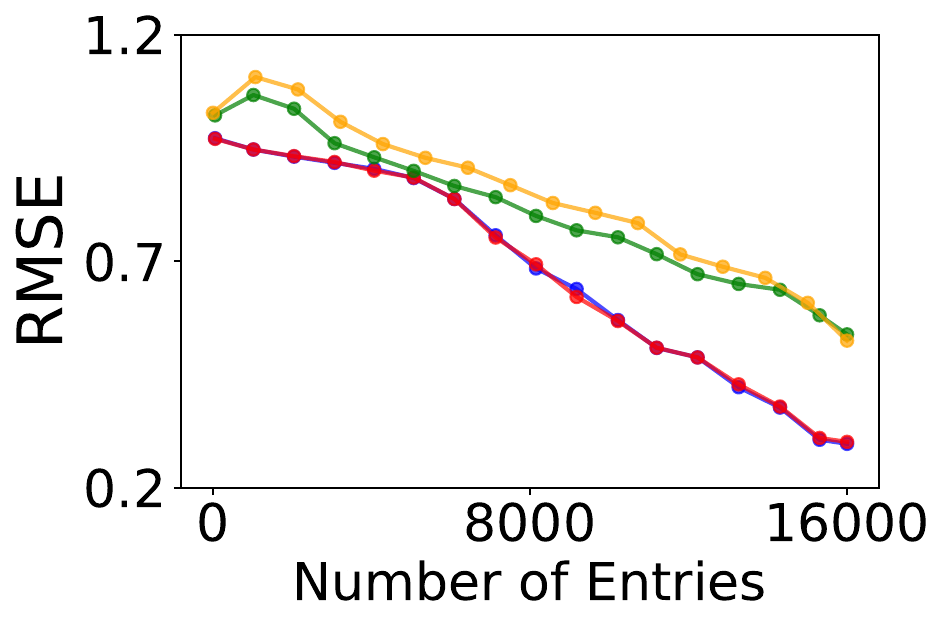}
			\caption{\textit{BeijingAir-2}}
		\end{subfigure}
		&
		\begin{subfigure}[t]{0.24\textwidth}
			\centering
			\includegraphics[width=\textwidth]{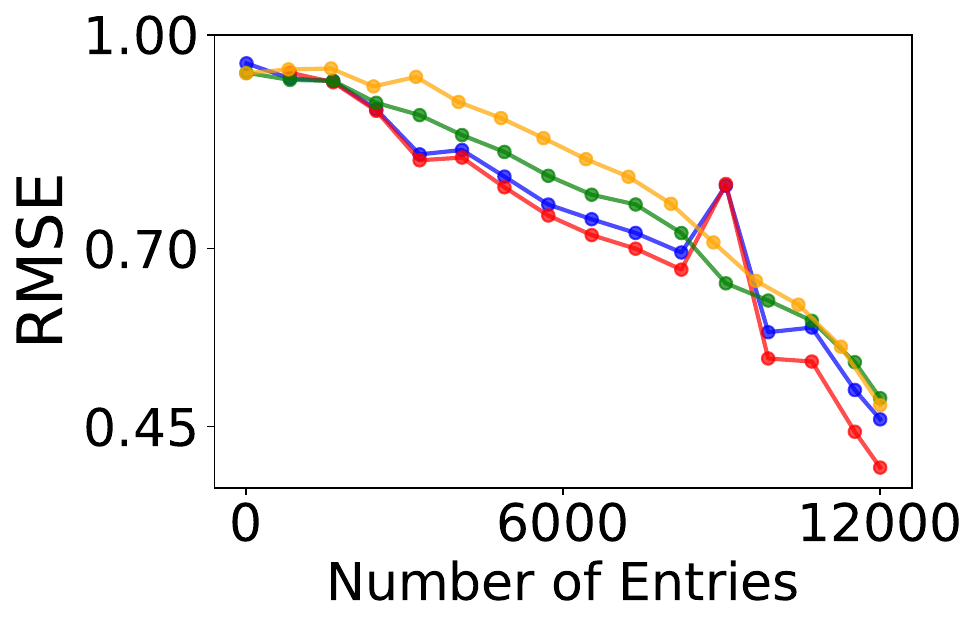}
			\caption{\textit{BeijingAir-3}}
		\end{subfigure}
	\end{tabular}	
%	\vspace{-0.1in}
	\caption{\small Online prediction error  with the number of processed  entries ($R=2$)} 	
	\label{fig:running-performance-r2}
%	\vspace{-0.2in}
\end{figure*}

\begin{figure*}[!ht]
	\centering
	\setlength\tabcolsep{0.01pt}
	\begin{tabular}[c]{cccc}
		%\multicolumn{4}{c}{\includegraphics[width=0.718\textwidth]{./figs_new/legend2.pdf}}\\
		\begin{subfigure}[t]{0.24\textwidth}
			\centering
			\includegraphics[width=\textwidth]{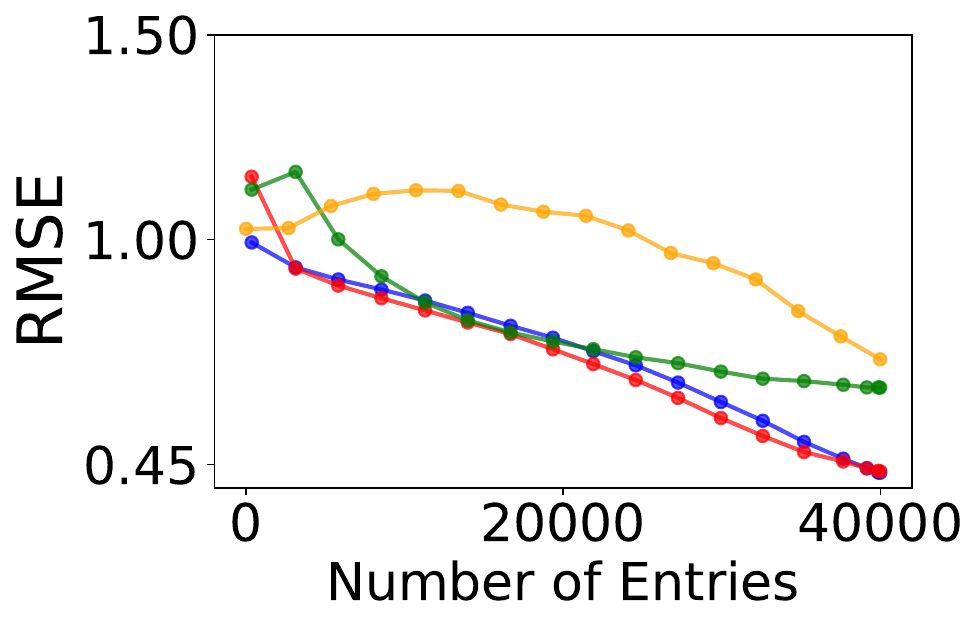}
			\caption{\textit{FitRecord}}
		\end{subfigure} 
		&
		\begin{subfigure}[t]{0.24\textwidth}
			\centering
			\includegraphics[width=\textwidth]{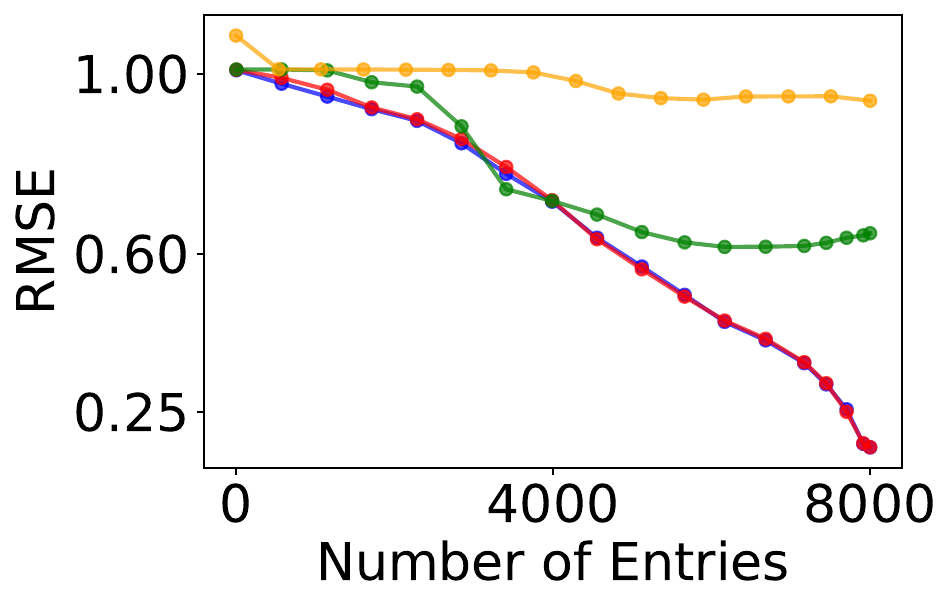}
			\caption{\textit{ServerRoom}}
		\end{subfigure} 
		&
		\begin{subfigure}[t]{0.24\textwidth}
			\centering
			\includegraphics[width=\textwidth]{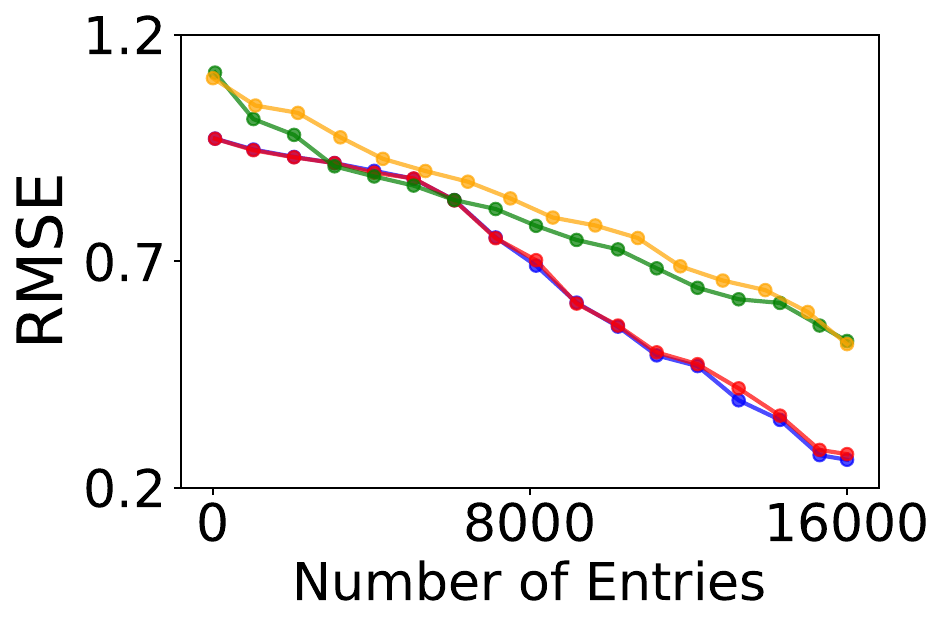}
			\caption{\textit{BeijingAir-2}}
		\end{subfigure}
		&
		\begin{subfigure}[t]{0.24\textwidth}
			\centering
			\includegraphics[width=\textwidth]{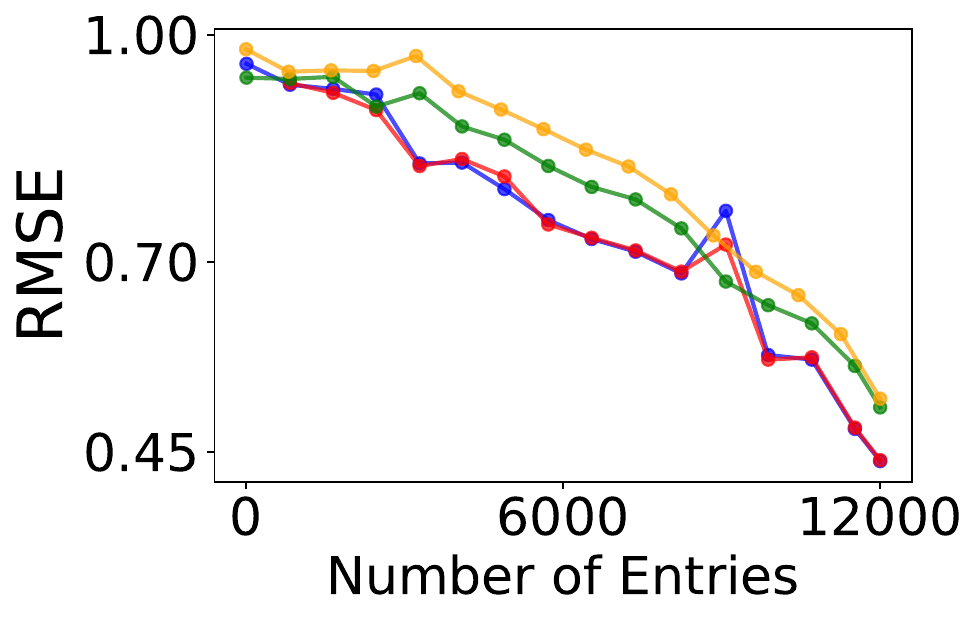}
			\caption{\textit{BeijingAir-3} }
		\end{subfigure}
	\end{tabular}	
%	\vspace{-0.1in}
	\caption{\small Online prediction error  with the number of processed  entries ($R=3$)} 	
	\label{fig:running-performance-r3}
	%\vspace{-0.2in}
\end{figure*}

\begin{figure*}[h]
	\centering
	\setlength\tabcolsep{0.01pt}
	\begin{tabular}[c]{cccc}
		%\multicolumn{4}{c}{\includegraphics[width=0.718\textwidth]{./figs_new/legend2.pdf}}\\
		\begin{subfigure}[t]{0.24\textwidth}
			\centering
			\includegraphics[width=\textwidth]{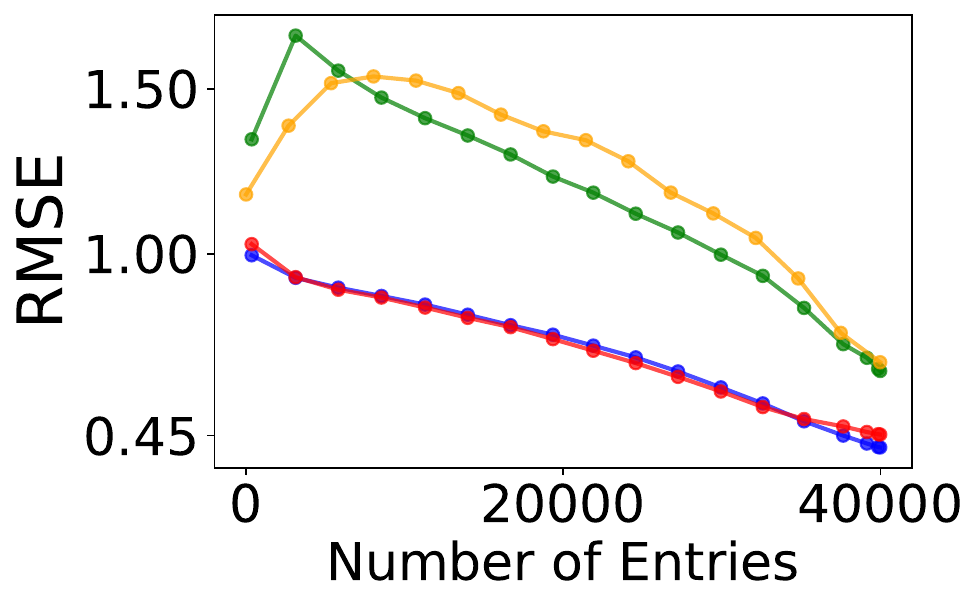}
			\caption{\textit{FitRecord} }
		\end{subfigure} 
		&
		\begin{subfigure}[t]{0.24\textwidth}
			\centering
			\includegraphics[width=\textwidth]{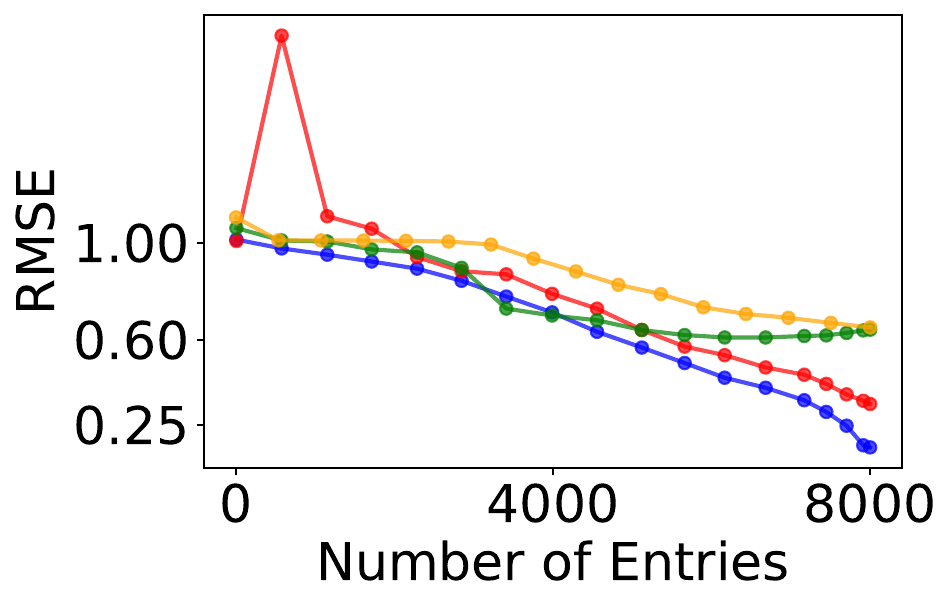}
			\caption{\textit{ServerRoom} }
		\end{subfigure} 
		&
		\begin{subfigure}[t]{0.24\textwidth}
			\centering
			\includegraphics[width=\textwidth]{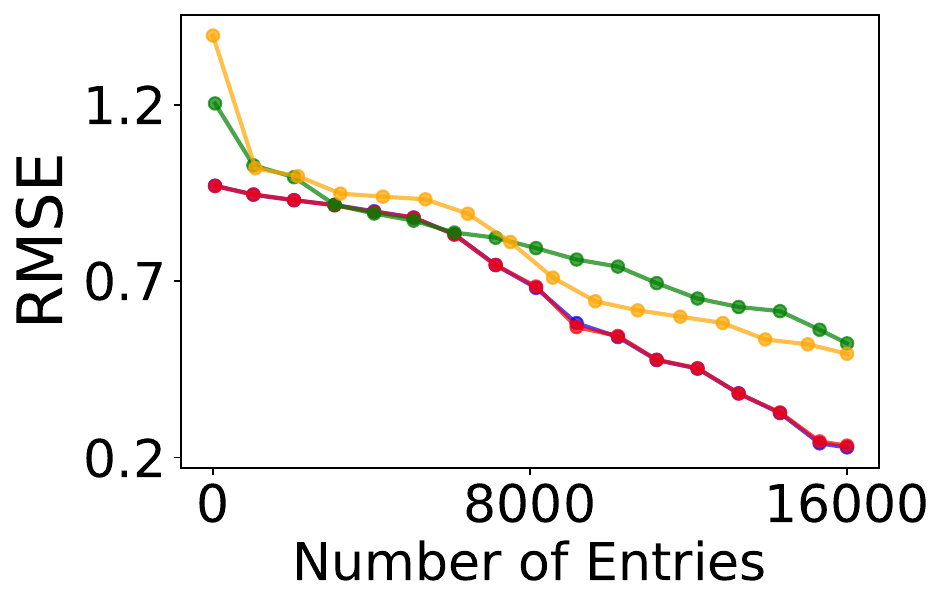}
			\caption{\textit{BeijingAir-2} }
		\end{subfigure}
		&
		\begin{subfigure}[t]{0.24\textwidth}
			\centering
			\includegraphics[width=\textwidth]{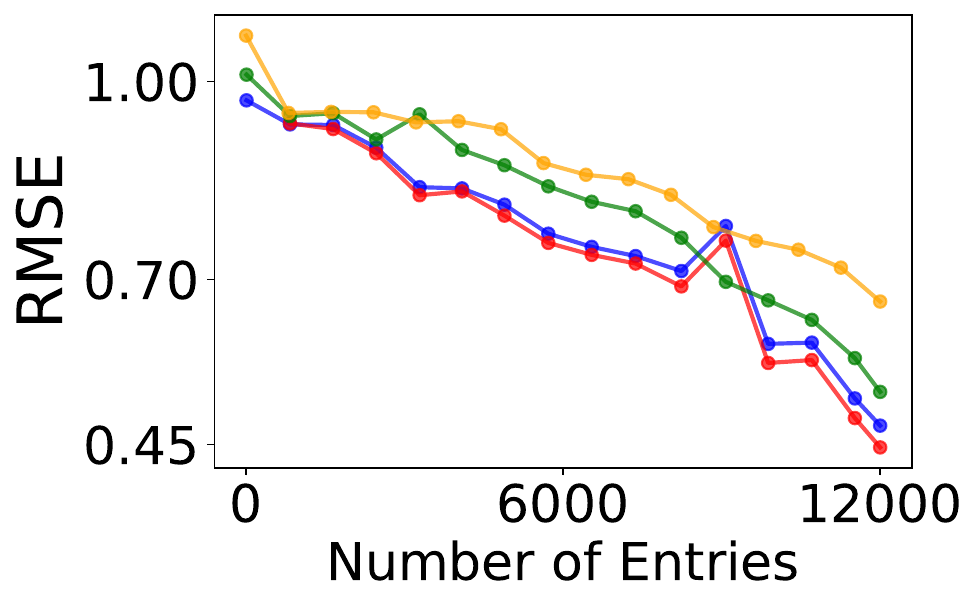}
			\caption{\textit{BeijingAir-3} }
		\end{subfigure}
	\end{tabular}	
%	\vspace{-0.1in}
	\caption{\small Online prediction error  with the number of processed  entries ($R=7$)} 	
	\label{fig:running-performance-r7}
%	\vspace{-0.2in}
\end{figure*}

We report for $R=2$, $R=3$ and $R=7$,  the final prediction error (after the data has been processed) of all the methods in Table \ref{table:Rank2res}, Table \ref{table:Rank3res}, and Table \ref{table:Rank7res}, respectively. 
We report for $R=2$, $R=3$ and $R=7$, the online predictive performance of the streaming decomposition approaches in Fig. \ref{fig:running-performance-r2}, Fig. \ref{fig:running-performance-r3}, and Fig.  \ref{fig:running-performance-r7}, respectively. 
\begin{table*}[]
	\centering
	\begin{small}
		\begin{tabular}{llcccc}
			\toprule
			& {RMSE}&  \textit{FitRecord} & \textit{ServerRoom} & \textit{BeijingAir-2} & \textit{BeijingAir-3}  \\ 
			\midrule
			\multirow{ 8}{*}{Static} &PTucker & $ 0.606\pm0.015 $ & $ 0.757\pm0.36 $  & $ 0.509\pm0.01 $  & $ 0.442\pm0.142 $ \\
			&Tucker-ALS& $ 0.914\pm0.01 $  & $ 0.991\pm0.016 $ & $ 0.586\pm0.016 $ & $ 0.896\pm0.032 $ \\
			&CP-ALS  & $ 0.926\pm0.013 $ & $ 0.997\pm0.016 $ & $ 0.647\pm0.041 $ & $ 0.918\pm0.031 $ \\
			&CT-CP   & $ 0.675\pm0.009 $ & $ 0.412\pm0.024 $ & $ 0.642\pm0.007 $ & $ 0.832\pm0.035 $ \\
			&CT-GP   & $ 0.611\pm0.009 $ & $ 0.218\pm0.021 $ & $ 0.723\pm0.01 $  & $ 0.88\pm0.026 $  \\
			&BCTT    & $ 0.604\pm0.019 $ & $ 0.715\pm0.352 $ & $ 0.504\pm0.01 $  & $ 0.799\pm0.027 $ \\
			&NONFAT  & $ 0.543\pm0.002 $ & $ \mathbf{0.132\pm0.002} $ & $ 0.425\pm0.002 $ & $ 0.878\pm0.014 $ \\
			&THIS-ODE& $ 0.544\pm0.005 $         & $ 0.142\pm0.004 $ & $ 0.553\pm0.015 $         & $ 0.876\pm0.027 $ \\
			\hline
			\multirow{5}{*}{Stream}&POST & $ 0.705\pm0.013 $ & $ 0.767\pm0.155 $ & $ 0.539\pm0.01 $  & $ 0.695\pm0.135 $ \\
			&ADF-CP  & $ 0.669\pm0.033 $ & $ 0.764\pm0.114 $ & $ 0.583\pm0.07 $  & $ 0.54\pm0.045 $  \\
			&BASS-Tucker        & $ 1\pm0.016 $     & $ 1\pm0.016 $     & $ 1.043\pm0.05 $  & $ 0.982\pm0.058 $ \\
			&SFTL-CP     & $ \mathbf{0.437\pm0.014} $ & $ {0.18\pm0.019} $  & $ \mathbf{0.323\pm0.019} $ & $ 0.462\pm0.009 $ \\
			&SFTL-Tucker  & $ 0.446\pm0.024 $ & $ 0.276\pm0.031 $ & $ 0.344\pm0.031 $ & $ \mathbf{0.417\pm0.035} $ \\ 
			\midrule
			& {MAE}    &           \\ 
			\midrule
			\multirow{ 8}{*}{Static}&PTucker  & $ 0.416\pm0.005 $ & $ 0.388\pm0.152 $ & $ 0.336\pm0.004 $ & $ 0.271\pm0.053 $ \\
			&Tucker-ALS & $ 0.676\pm0.008 $ & $ 0.744\pm0.01 $  & $ 0.408\pm0.008 $ & $ 0.669\pm0.02 $  \\
			&CP-ALS   & $ 0.686\pm0.011 $ & $ 0.748\pm0.009 $ & $ 0.454\pm0.057 $ & $ 0.691\pm0.016 $ \\
			&CT-CP          & $ 0.466\pm0.005 $ & $ 0.295\pm0.029 $ & $ 0.49\pm0.006 $  & $ 0.642\pm0.02 $  \\
			&CT-GP         & $ 0.424\pm0.006 $ & $ 0.155\pm0.012 $ & $ 0.517\pm0.01 $  & $ 0.626\pm0.01 $  \\
			&BCTT          & $ 0.419\pm0.015 $ & $ 0.534\pm0.263 $ & $ 0.343\pm0.003 $ & $ 0.579\pm0.018 $ \\
			&NONFAT      & $ 0.373\pm0.001 $ & $ \mathbf{0.083\pm0.001} $ & $ 0.282\pm0.002 $ & $ 0.622\pm0.006 $ \\
			&THIS-ODE    & $ 0.377\pm0.003 $         & $ 0.097\pm0.003 $ & $ 0.355\pm0.008 $ & $ 0.606\pm0.015 $ \\
			\hline
			\multirow{5}{*}{Stream}&POST        & $ 0.485\pm0.008 $ & $ 0.564\pm0.091 $ & $ 0.368\pm0.008 $ & $ 0.517\pm0.123 $ \\
			&ADF-CP    & $ 0.462\pm0.022 $ & $ 0.574\pm0.073 $ & $ 0.401\pm0.029 $ & $ 0.415\pm0.038 $ \\
			&BASS      & $ 0.777\pm0.039 $ & $ 0.749\pm0.01 $  & $ 0.871\pm0.125 $ & $ 0.727\pm0.029 $ \\
			&SFTL-CP   & $ \mathbf{0.248\pm0.005} $ & $ 0.126\pm0.007 $ & $ \mathbf{0.199\pm0.005} $ & $ 0.311\pm0.004 $ \\
			&SFTL-Tucker & $ 0.25\pm0.01 $   & $ 0.203\pm0.032 $ & $ 0.218\pm0.02 $  & $ \mathbf{0.261\pm0.023} $ \\
			\bottomrule
		\end{tabular}
	\end{small}
	%\vspace{-0.1in}
	\caption{\small Final prediction error \cmt{of the static and streaming decomposition methods,} with $R=2$. The results were averaged from five runs.}
	\label{table:Rank2res}
	%\vspace{-0.1in}
\end{table*}

\begin{table*}[]
	\centering
	\begin{small}
		\begin{tabular}{llcccc}
			\toprule
			& {RMSE}&  \textit{FitRecord} & \textit{ServerRoom} & \textit{BeijingAir-2} & \textit{BeijingAir-3}  \\ 
			\midrule
			\multirow{ 8}{*}{Static}&PTucker & $ 0.603\pm0.045 $ & $ 0.677\pm0.129 $ & $ 0.464\pm0.012 $ & $ 0.421\pm0.074 $ \\
			&Tucker-ALS& $ 0.885\pm0.007 $ & $ 0.989\pm0.014 $ & $ 0.559\pm0.017 $ & $ 0.863\pm0.032 $ \\
			&CP-ALS & $ 0.907\pm0.015 $ & $ 0.993\pm0.016 $ & $ 0.594\pm0.031 $ & $ 0.901\pm0.03 $ \\
			&CT-CP & $ 0.666\pm0.008 $ & $ 0.5\pm0.2 $     & $ 0.641\pm0.006 $ & $ 0.819\pm0.019 $ \\
			&CT-GP      & $ 0.606\pm0.008 $ & $ 0.217\pm0.025 $ & $ 0.749\pm0.014 $ & $ 0.895\pm0.054 $ \\
			&BCTT       & $ 0.576\pm0.015 $ & $ 0.358\pm0.082 $ & $ 0.454\pm0.011 $ & $ 0.829\pm0.028 $ \\
			&NONFAT    & $ 0.517\pm0.002 $ & $ \mathbf{0.129\pm0.002} $ & $ 0.408\pm0.005 $ & $ 0.877\pm0.014 $ \\ 
			&THIS-ODE   & $ 0.528\pm0.005 $         & $ 0.132\pm0.002 $ & $ 0.544\pm0.014 $         & $ 0.878\pm0.026 $ \\
			\hline
			\multirow{5}{*}{Stream}&POST    & $ 0.706\pm0.034 $ & $ 0.741\pm0.161 $ & $ 0.518\pm0.016 $ & $ 0.622\pm0.123 $ \\
			&ADF-CP     & $ 0.641\pm0.009 $ & $ 0.652\pm0.012 $ & $ 0.542\pm0.012 $ & $ 0.518\pm0.003 $ \\
			&BASS-Tucker       & $ 1.008\pm0.017 $ & $ 1\pm0.016 $     & $ 1.035\pm0.038 $ & $ 0.99\pm0.034 $  \\
			&SFTL-CP    & $ 0.434\pm0.014 $ & $ 0.178\pm0.006 $ & $ \mathbf{0.288\pm0.017} $ & $ 0.454\pm0.011 $ \\
			&SFTL-Tucker   & $ \mathbf{0.418\pm0.01} $  & $ 0.289\pm0.096 $ & $ 0.314\pm0.049 $ & $ \mathbf{0.41\pm0.013\mathbf} $  \\ 
			\midrule
			& {MAE}    &           \\ 
			\midrule
			\multirow{ 8}{*}{Static}&PTucker & $ 0.392\pm0.009 $ & $ 0.323\pm0.053 $ & $ 0.307\pm0.005 $ & $ 0.197\pm0.029 $ \\
			&Tucker-ALS  & $ 0.648\pm0.012 $ & $ 0.743\pm0.008 $ & $ 0.39\pm0.008 $  & $ 0.651\pm0.018 $ \\
			&CP-ALS      & $ 0.666\pm0.013 $ & $ 0.746\pm0.01 $  & $ 0.415\pm0.022 $ & $ 0.676\pm0.021 $ \\
			&CT-CP    	 & $ 0.462\pm0.005 $ & $ 0.348\pm0.141 $ & $ 0.489\pm0.006 $ & $ 0.632\pm0.015 $ \\
			&CT-GP    & $ 0.419\pm0.005 $ & $ 0.158\pm0.022 $ & $ 0.544\pm0.012 $ & $ 0.627\pm0.015 $ \\
			&BCTT     & $ 0.392\pm0.004 $ & $ 0.267\pm0.067 $ & $ 0.299\pm0.006 $ & $ 0.607\pm0.027 $ \\
			&NONFAT   & $ 0.355\pm0.001 $ & $ \mathbf{0.078\pm0.001} $ & $ 0.265\pm0.003 $ & $ 0.622\pm0.006 $ \\
			&THIS-ODE & $ 0.363\pm0.004 $         & $ 0.083\pm0.002 $ & $ 0.348\pm0.006 $         & $ 0.603\pm0.009 $ \\
			\hline
			\multirow{5}{*}{Stream}&POST   & $ 0.482\pm0.022 $ & $ 0.54\pm0.102 $  & $ 0.351\pm0.009 $ & $ 0.442\pm0.109 $ \\
			&ADF-CP   & $ 0.445\pm0.006 $ & $ 0.5\pm0.009 $   & $ 0.381\pm0.006 $ & $ 0.393\pm0.009 $ \\
			&BASS      & $ 0.822\pm0.024 $ & $ 0.749\pm0.009 $ & $ 0.919\pm0.041 $ & $ 0.73\pm0.018 $  \\
			&SFTL-CP  & $ 0.246\pm0.005 $ & $ 0.121\pm0.003 $ & $ \mathbf{0.176\pm0.006} $ & $ 0.305\pm0.006 $ \\
			&SFTL-Tucker &$ \mathbf{ 0.24\pm0.002} $  & $ 0.18\pm0.042 $  & $ 0.196\pm0.03 $  & $ \mathbf{0.263\pm0.011} $ \\
			\bottomrule
		\end{tabular}
	\end{small}
	%\vspace{-0.1in}
	\caption{\small Final prediction error \cmt{of the static and streaming decomposition methods,} with $R=3$. The results were averaged from five runs.}
	\label{table:Rank3res}
	%\vspace{-0.1in}
\end{table*}

\begin{table*}[]
	\centering
	\begin{small}
		\begin{tabular}{llcccc}
			\toprule
			& {RMSE}&  \textit{FitRecord} & \textit{ServerRoom} & \textit{BeijingAir-2} & \textit{BeijingAir-3}  \\ 
			\midrule
			\multirow{ 8}{*}{Static}&PTucker & $ 0.603\pm0.045 $ & $ 0.677\pm0.129 $ & $ 0.464\pm0.012 $ & $ 0.421\pm0.074 $ \\
			&Tucker-ALS& $ 0.826\pm0.003 $ & $ 0.983\pm0.016 $ & $ 0.586\pm0.018 $ & $ 0.825\pm0.026 $ \\
			&CP-ALS & $ 0.878\pm0.012 $ & $ 0.994\pm0.013 $ & $ 0.897\pm0.215 $ & $ 0.863\pm0.024 $\\
			&CT-CP   & $ 0.663\pm0.008 $ & $ 0.384\pm0.008 $ & $ 0.64\pm0.007 $  & $ 0.818\pm0.019 $ \\
			&CT-GP   & $ 0.603\pm0.006 $ & $ 0.381\pm0.303 $ & $ 0.766\pm0.016 $ & $ 0.904\pm0.046 $ \\
			&BCTT     & $ 0.498\pm0.011 $ & $ 0.194\pm0.017 $ & $ 0.368\pm0.01 $  & $ 0.813\pm0.028 $ \\
			&NONFAT   & $ 0.497\pm0.003 $ & $ \mathbf{0.128\pm0.002} $ & $ 0.394\pm0.004 $ & $ 0.88\pm0.013 $  \\
			&THIS-ODE  & $ 0.138\pm0.003 $ & $ 0.554\pm0.016 $         & $ 0.878\pm0.027 $ \\
			\hline
			\multirow{5}{*}{Stream}&POST & $ 0.675\pm0.012 $ & $ 0.707\pm0.14 $  & $ 0.519\pm0.017 $ & $ 0.738\pm0.068 $ \\
			&ADF-CP  & $ 0.652\pm0.01 $  & $ 0.646\pm0.008 $ & $ 0.548\pm0.012 $ & $ 0.552\pm0.026 $ \\
			&BASS-Tucker & $ 0.604\pm0.043 $ & $ 0.493\pm0.071 $ & $ 0.391\pm0.005 $ & $ 0.634\pm0.083 $ \\
			&SFTL-CP & $ \mathbf{0.424\pm0.006} $ & $ 0.166\pm0.013 $ & $ 0.256\pm0.013 $ & $ 0.481\pm0.006 $ \\
			&SFTL-Tucker   & $ 0.448\pm0.009 $ & $ 0.406\pm0.052 $ & $ \mathbf{0.249\pm0.017} $ & $ \mathbf{0.432\pm0.019} $ \\ 
			\midrule
			& {MAE}    &           \\ 
			\midrule
			\multirow{ 8}{*}{Static}&PTucker & $ 0.353\pm0.005 $ & $ 0.305\pm0.042 $ & $ 0.248\pm0.004 $ & $ 0.32\pm0.038 $  \\
			&Tucker-ALS & $ 0.6\pm0.002 $   & $ 0.737\pm0.009 $ & $ 0.392\pm0.011 $ & $ 0.619\pm0.015 $ \\
			&CP-ALS     & $ 0.64\pm0.009 $  & $ 0.745\pm0.008 $ & $ 0.593\pm0.121 $ & $ 0.637\pm0.015 $ \\
			&CT-CP    	 & $ 0.459\pm0.005 $ & $ 0.27\pm0.003 $  & $ 0.488\pm0.005 $ & $ 0.626\pm0.012 $ \\
			&CT-GP   & $ 0.412\pm0.004 $ & $ 0.282\pm0.23 $  & $ 0.557\pm0.009 $ & $ 0.628\pm0.01 $  \\
			&BCTT   & $ 0.342\pm0.005 $ & $ 0.157\pm0.015 $ & $ 0.234\pm0.005 $ & $ 0.581\pm0.022 $ \\
			&NONFAT   & $ 0.335\pm0.002 $ & $ \mathbf{0.077\pm0.002} $ & $ 0.256\pm0.003 $ & $ 0.627\pm0.005 $ \\
			&THIS-ODE & $ 0.362\pm0.002 $         & $ 0.089\pm0.002 $ & $ 0.357\pm0.007 $         & $ 0.603\pm0.013 $ \\
			\hline
			\multirow{5}{*}{Stream}&POST & $ 0.461\pm0.008 $ & $ 0.518\pm0.087 $ & $ 0.357\pm0.011 $ & $ 0.558\pm0.058 $ \\
			&ADF-CP   & $ 0.451\pm0.006 $ & $ 0.489\pm0.009 $ & $ 0.384\pm0.014 $ & $ 0.411\pm0.025 $ \\
			&BASS  & $ 0.745\pm0.026 $ & $ 0.749\pm0.01 $  & $ 0.903\pm0.044 $ & $ 0.721\pm0.038 $ \\
			&SFTL-CP  & $ \mathbf{0.243\pm0.003} $ & $ 0.111\pm0.008 $ & $ 0.159\pm0.004 $ & $ 0.323\pm0.003 $ \\
			&SFTL-Tucker  & $ 0.253\pm0.004 $ & $ 0.273\pm0.033 $ & $ \mathbf{0.144\pm0.008} $ & $ \mathbf{0.273\pm0.016} $ \\
			\bottomrule
		\end{tabular}
	\end{small}
	%\vspace{-0.1in}
	\caption{\small Final prediction error \cmt{of the static and streaming decomposition methods,} with $R=7$. The results were averaged from five runs.}
	\label{table:Rank7res}
	%\vspace{-0.1in}
\end{table*}

\section{Running Time}\label{sect:running-time}
 As compared with static (non-streaming) methods, such as BCTT, our method is faster and more efficient. That is because whenever new data comes in, the static methods have to retrain the model from scratch and iteratively access the whole data accumulated so far, while our method only performs incremental updates and never needs to revisit the past data. To demonstrate this point, we compared the training time of our method with BCTT on \textit{BeijingAir2} dataset. All the methods were run on a Linux workstation. From Table \ref{table:Running-Time}, we can see a large speed-up of our method with both the CP and Tucker form. The higher the rank ($R$), the more significant the speed-up.
\begin{table*}[]
	\centering
	\begin{small}
		\begin{tabular}{lcccc}
			\toprule
			{ } & {  {$R=2$}} & {  {$R=3$}} & {  {$R=5$}} & {  {$R=7$}} \\
			\hline
			{  \ours-CP}                      & {  27.1}         & {  27.2}         & {  28.5}         & {  29.1}         \\
			{  \ours-Tucker}                  & {  32.3}         & {  35.6}         & {  43.2}         & {  59.3}         \\
			{  BCTT}                         & {  49.5}         & {  56.1}         & {  72.1}         & {  136.7}       \\
			\bottomrule
		\end{tabular}
	\end{small}
	%\vspace{-0.1in}
	\caption{\small Running time in seconds on \textit{BeijingAir2} dataset.}
	\label{table:Running-Time}
\end{table*}